\documentclass[utf8]{frontiersSCNS}

\usepackage{url,hyperref,lineno,microtype,subcaption}
\usepackage[onehalfspacing]{setspace}

\usepackage{amsmath}
\usepackage{amssymb}

\usepackage{hyperref}

\usepackage{graphicx}

\usepackage{fancyvrb}
\DefineVerbatimEnvironment{Sinput}{Verbatim}{fontshape=sl}
\DefineVerbatimEnvironment{Soutput}{Verbatim}{}
\DefineVerbatimEnvironment{Scode}{Verbatim}{fontshape=sl}

\DefineVerbatimEnvironment{Code}{Verbatim}{}
\DefineVerbatimEnvironment{CodeInput}{Verbatim}{fontshape=sl}
\DefineVerbatimEnvironment{CodeOutput}{Verbatim}{}

\usepackage{cite}

\usepackage{color}

\def\keyFont{\fontsize{8}{11}\helveticabold }

\usepackage{booktabs}
\usepackage{longtable}
\usepackage{array}
\usepackage{multirow}
\usepackage{wrapfig}
\usepackage{float}
\usepackage{colortbl}
\usepackage{pdflscape}
\usepackage{tabu}
\usepackage{threeparttable}
\usepackage{threeparttablex}
\usepackage[normalem]{ulem}
\usepackage{makecell}
\usepackage{xcolor}

\def\Authors{
  Jayden L. Macklin-Cordes\,\textsuperscript{1*},
  Erich R. Round\,\textsuperscript{1,2,3}}

\def\Address{

  \textsuperscript{1} School of Languages and Cultures, The University of Queensland,  Brisbane,  QLD,  Australia
  
  \textsuperscript{2} Department of Linguistic and Cultural Evolution, Max Planck Institute for the Science of Human History,  Jena,   Germany
  
  \textsuperscript{3} Surrey Morphology Group, University of Surrey,  Guildford,   United Kingdom
  }

  \def\corrAuthor{Jayden L. Macklin-Cordes}\def\corrEmail{\href{mailto:j.macklincordes@uq.edu.au}{\nolinkurl{j.macklincordes@uq.edu.au}}}
  \def\firstAuthorLast{MACKLIN-CORDES {et~al.}}

\begin{document}
\onecolumn
\firstpage{1}

\title[Re-evaluating phoneme frequencies]{Re-evaluating phoneme frequencies}
\author[\firstAuthorLast]{\Authors}
\address{} %This field will be automatically populated
\correspondance{} %This field will be automatically populated

\extraAuth{}% If there are more than 1 corresponding author, comment this line and uncomment the next one.
%\extraAuth{corresponding Author2 \\ Laboratory X2, Institute X2, Department X2, Organization X2, Street X2, City X2 , State XX2 (only USA, Canada and Australia), Zip Code2, X2 Country X2, email2@uni2.edu}

\maketitle

\begin{abstract}

\noindent
Causal processes can give rise to distinctive distributions in the linguistic variables that they affect. Consequently, a secure understanding of a variable's distribution can hold a key to understanding the forces that have causally shaped it. A storied distribution in linguistics has been Zipf's law, a kind of power law. In the wake of a major debate in the sciences around power-law hypotheses and the unreliability of earlier methods of evaluating them, here we re-evaluate the distributions claimed to characterize phoneme frequencies. We infer the fit of power laws and three alternative distributions to 166 Australian languages, using a maximum likelihood framework. We find evidence supporting earlier results, but also nuancing them and increasing our understanding of them. Most notably, phonemic inventories appear to have a Zipfian-like frequency structure among their most-frequent members (though perhaps also a lognormal structure) but a geometric (or exponential) structure among the least-frequent. We compare these new insights the kinds of causal processes that affect the evolution of phonemic inventories over time, and identify a potential account for why, despite there being an important role for phonetic substance in phonemic change, we could still expect inventories with highly diverse phonetic content to share similar distributions of phoneme frequencies. We conclude with priorities for future work in this promising program of research.

\tiny
 \keyFont{ \section{Keywords:} Power laws; Zipf's law; phoneme inventories; distributions; maximum likelihood; Australian languages; phonology } 

\end{abstract}

\hypertarget{introduction}{%
\section*{Introduction}\label{introduction}}
\addcontentsline{toc}{section}{Introduction}

Linguistic theorists seek to reveal causal mechanisms which explain the observable diversity of human language. Good causal hypotheses are often suggested by the mathematical distribution that a linguistic variable is described by, owing to the fact that the distribution can be understood as an emergent outcome of some underlying causal process, and that a given mathematical distribution will be consistent with only certain mathematical kinds of underlying processes. Consequently, it is important for the development of theory that proposed claims about distributions be as sound as possible. For instance, one of the most famous distributions in linguistics is the Zipfian distribution, which technically speaking is a kind of power law. Recently, however, the evaluation of putative power laws across the sciences have come under intense scrutiny and often been found wanting. In response, methodologists have developed more rigorous and secure methods for diagnosing power laws and for distinguishing them from similar but significantly different distributions. For linguists, this creates an opportunity, to re-examine our own putative power law distributions, and by doing so to improve the pathway to sound explanatory theorizing.

Here, we re-evaluate the status of mathematical distributions for characterizing phoneme frequencies. Previous studies have proposed that phoneme frequencies follow a particular member of the power law family, the Yule-Simon distribution (Martindale et al., 1996; Tambovtsev and Martindale, 2007). But in the wake of a recent, major debate across the sciences regarding power laws, a reconsideration of this earlier research is timely. In this paper, we apply state-of-the-art maximum likelihood methods for the detection and assessment of power laws, to derive a better understanding of the distributions that do and do not describe phoneme frequencies well. Our results clear the way for more informed research into the ultimate, processual causes behind the frequency patterns of phonemes in human language.

Our focus here is on distributions and their potential for shedding light on language. Distributions are properties of variables. A variable can be defined as the set of values that characterize something, be it a \emph{sample} (e.g.~a set of languages), or a \emph{real population} that the sample is drawn from (e.g.~the set of all current languages), or even an \emph{idealized population} which the real population is believed to approximate (e.g.~the set of all possible languages). Often, the ultimate object of scientific interest is an idealized population, and thus its distribution. In empirical work, we cannot directly access this ultimate object of interest, and so we rely on real populations, or very often, a sample. Consequently, although we may have direct access only to the literal distribution of a sample, with its many idiosyncrasies, we tend to be more concerned with an overall pattern which we believe it approximates, one which often is elegantly characterized by a distribution which mathematically is relatively simple.

With these motivations in mind, how can an investigator decide that a certain distribution characterizes a variable satisfactorily? One method is visual inspection. This typically involves observing a close match between two histogram plots: one of the data and one of the candidate distribution, or plotting a regression of the data against the candidate distribution. In work on phoneme frequencies, visual inspection has been the primary method of assessing candidate distributions. A more rigorous alternative is to apply quantitative, statistical tests to evaluate how well a particular distribution model (such as the normal distribution) fits a data sample. The purpose of these tests is not to prove definitively that some variable follows a particular distribution, but rather to quantify the degree to which a sample's distribution is consistent with its having been drawn from a population of a particular distribution.\footnote{Within frequentist statistics, there are tests which test against the null hypothesis that data are drawn from a particular distribution (Shapiro-Wilk, Kolmogorov-Smirnov, Pearson's chi-squared test, to name a few). Within a Bayesian framework (Spiegelhalter, 1980; Farrell and Rogers-Stewart, 2006), the approach is to calculate the likelihood of observing the data given a distribution and a set of the distribution's parameters. Bayes factors can be computed to compare the relative likelihoods of observing the data given competing kinds of distributions and parameter sets.} Some of these tests will be the subject of the sections that follow.

A strong motivation for testing the consistency of observed data with a particular distribution, is that many distributions can be described as the \emph{outcome} of certain kinds of processes (Frank, 2014). Thus there is a direct link between the quality of our evaluation of distributions, and the reasonableness of the causal, explanatory hypotheses we subsequently entertain. Consider for example a so-called \emph{preferential attachment process}, also known as a Yule process or rich-get-richer process. This can be imagined as having a set of urns, into which balls are added one at a time. Specifically, the urn to which a new ball is added is selected with a probability proportional to the number of balls already in the urn. This simple process has an interesting outcome. Initially, each urn is equally likely to be selected, but the distribution will soon skew, as urns with more balls accumulate additional balls faster than the others. If we \emph{rank} the urns in terms of which has the most balls, then with time, the relationship between an urn's rank and how many balls it contains will come to obey a \emph{power law}. A power law is a mathematical a relationship between two quantities where one varies as a power of the other. Consequently, if a variable can be shown to be consistent with a power law distribution, then this is consistent with there existing a preferential attachment process as the causal mechanism underlying the behaviour of variable. Yule (1925; see also Albert et al., 2011) made this connection a century ago. Yule showed that among flowering plants the level of species richness within a genus follows a power law, and linked that observation to a preferential attachment mechanism.

Since Yule's first demonstration of the link between power law distributions and preferential attachment processes, power laws have been used to characterize the distributions of a diverse array of phenomena in the natural and physical world and in human society (Clauset et al., 2009, 661). City populations (Gabaix, 1999; Levy, 2009; Malevergne et al., 2011), authorship of scientific publications (Simon, 1955), income distribution (Simon, 1955), the superstar phenomenon in the music industry (Chung and Cox, 1994) and the network topology of the Internet (Faloutsos et al., 1999) are but a few of the phenomena for which power laws have been proposed (see Newman (2005), pp.~327--329 for further examples). And in linguistics, the Zipfian distribution has been used to characterize word frequencies in text corpora (Estoup, 1916; Zipf, 1932, 1949).

Given the apparent pervasiveness of power laws in a diverse range of unrelated contexts, it is little surprise that there is a rich vein of literature dedicated to evaluating power laws (Clauset et al., 2009, 662) as well as a century of theorizing on the mechanistic processes by which they arise (see Newman, 2005, 336--348; Mitzenmacher, 2004, 230--243). However, verifying the presence of a power law is not a straightforward task (Stumpf and Porter, 2012, 666), and validation of earlier power law proposals using increasingly robust and powerful statistical methods is an active line of inquiry across many fields of science (Malevergne et al., 2011).\footnote{The question of whether certain phenomena are characterized best by a power law model or some other distribution can be contentious. See, for example, the debate between Eeckhout (2004) and Levy (2009) on the distribution of city population sizes (the former favouring a lognormal model, the latter favouring a power law). Another example concerns the distribution of computer file sizes, where Barford and Crovella (1998) and Barford et al. (1999) argue in favour of a power law model and Downey (2001) argues in favour of a lognormal model.}

The traditional approach to power law validation, following Pareto's (1897) work on wealth distribution, was visual inspection. When visually inspecting a histogram plotted on log-log scales, a straight line would suggest the presence of a power law. The defining shape parameter (see below), \(\alpha\), could then be obtained by calculating the slope of the straight line using standard linear regression (Clauset et al., 2009, 665; Urzúa, 2011, 254) and the \(R^2\) statistic could give an indication of the goodness of fit of the model. However, it has since been demonstrated that this traditional approach can be systematically unreliable (Clauset et al., 2009, 665). A key assumption when estimating the standard error of the slope (shaded in grey) is that noise in the data is normally distributed, however, this is not the case for the logarithms of frequency data (Clauset et al., 2009, 691). Further, the \(R^2\) statistic commonly used to validate the presence of a power law (including by Tambovtsev and Martindale, 2007), has low statistical power. That is, it often fails to distinguish between data truly drawn from a power law distribution and data drawn from other distribution types. This unreliability becomes particularly acute when there is a small number of observations, since the ability to distinguish a power law distribution from other similar distributions, including the log-normal, using \(R^2\) is reduced (Clauset et al., 2009, 691).

To remedy the shortcomings of earlier methods, Clauset et al. (2009) developed power law validation procedures within a more rigorous, maximum likelihood framework. These procedures have since been adopted widely in the literature (for example, Touboul and Destexhe, 2010; Cho et al., 2011; Brzezinski, 2014; and Lee and Kim, 2018), but have not previously been applied to phoneme frequencies.

Some brief mathematical preliminaries will be useful here. When we refer to distributions, we are referring to mathematically defined \emph{functions}, that relate one quantity to another. Those functions may in addition have \emph{free parameters} which can be varied in order to produce a family of closely related distributions. A power law is a relationship between two quantities where one varies as a fixed power of the other, for example \(y = x^3\), or \(y = x^{-2}\) (which can also be written \(y = 1/x^2\)). For present purposes, where we will not be concerned with negative quantities or zeros, we will use a more narrow definition by Clauset et al. (2009, 662), who define a power law as a relationship in which a quantity, \(x\), is drawn from the distribution defined in Equation \eqref{eq:power-law}, where the free parameter \(\alpha\) is greater than zero and the variable \(x\) likewise is greater than zero.\footnote{For the area under the distribution curve to integrate properly to 1, the power function \(1/x^{\alpha}\) must be multiplied by a normalization constant (denoted \(C\) in the probability density function \(p(x) = C/x^{\alpha}\)). The normalization constant will be calculated differently depending on the value of \(\alpha\) and whether \(x\) is continuous or discrete. Clauset et al. (2009, 664) give some examples.} (The symbol `\(\propto\)' means `is proportional to'.) For example, \(x\) might denote items' frequencies, while \(p(x)\) is the probability that a given item has a frequency of \(x\).

\begin{equation}
p(x) \propto \frac{1}{x^{\alpha}},\quad x > 0
\label{eq:power-law}
\end{equation}

In practice, Clauset et al. (2009, 662) observe that the exponent (or `scaling parameter'), \(\alpha\), typically, though not exclusively, falls in the range \(2 < \alpha < 3\). They also observe that, in practice, many phenomena will not actually obey a power law for all values of \(x\). Rather, the power law will apply to values only above some minimum threshold value, \(x_{min}\). For example, in frequency data, it may be that only items whose frequencies meet or exceed a lower threshold will follow a power law. More generally, power law distributions come in a variety of specific forms, with different numbers of free parameters. We detail some of these in greater depth in the Discussion section below.

A distinction can be made between power laws that apply to continuous variables and those that apply to discrete ones. Frequency data, including the phoneme frequencies used in this study, are typically discrete. Zipf's Law \eqref{eq:zipfs-law} applies to a discrete number of \(n\) observations whose values, \(x\), are ranked by descending magnitude \(x_1 \geq x_2 \geq \ldots \geq x_n\). For example, \(x\) may be the token frequency of \(n\) types, with \(x_k\) the frequency of the \(k\)th-ranked type. In \eqref{eq:zipfs-law}, the quantity \(p(x_k)\) is the relative frequency of the \(k\)th-ranked type (i.e., its frequency scaled such that the \(n\) relative frequencies sum to 1).\footnote{This is equivalent to the probability that a token selected at random belongs to the \(k\)th-ranked type.}

\begin{equation}
p(x_k) \propto \frac{1}{k^{\alpha}},\quad 1 \leq k \leq n
\label{eq:zipfs-law}
\end{equation}

There is a long history of studying power laws in linguistics, however, the evaluation of statistical support for a power law relationship is far from straightforward and remains topical across a wide range of scientific fields (Stumpf and Porter, 2012). Although several different models have been compared for their goodness-of-fit to the frequencies of phonological segments (Martindale et al., 1996; Tambovtsev and Martindale, 2007), the method used to measure fit (using the \(R^2\) statistic) has been shown to be systematically unreliable (Clauset et al., 2009). Accordingly, the methodological limitations of previous studies and the renewed, general scientific interest in power law phenomena motivate the re-evaluation of a power law model with respect to phoneme frequencies. Our goal here is to verify the presence or absence of power law behaviour in the frequency distributions of phonological segments in the lexicons of Australian languages. It is, to the best of our knowledge, the first attempt to validate a power law model for phonological segments using a maximum likelihood framework as suggested by Clauset et al. (2009).

\hypertarget{materials-and-methods}{%
\section*{Materials and methods}\label{materials-and-methods}}
\addcontentsline{toc}{section}{Materials and methods}

\hypertarget{data}{%
\subsection*{Data}\label{data}}
\addcontentsline{toc}{subsection}{Data}

As our data, we take phoneme frequencies in the lexicons of 166 language varieties of Australia, covering the 19 of Australia's language families for which phonemic lexical data is available, including all major branches of the Pama-Nyungan family which dominates the continent.\footnote{The dataset closely approximates an exhaustive sample, containing around 80\% of Australian languages for which phonemic lexical data is known to exist, and on the order of 40\% of the varieties that were spoken at the onset of European colonization.} Our choice of dataset brings advantages and limitations. Prior studies of phoneme frequency distributions have overwhelming focused on languages of Eurasia, and it is valuable to confirm whether similar results emerge on other continents. Because Australian languages are known to have quite similar phonemic inventories across the continent (Capell, 1956; Dixon, 1980; Busby, 1982; Hamilton, 1996; Baker, 2014; Round, 2019, 2020), it is useful to confirm whether, even under conditions of highly constrained variation in the substance of the phoneme inventories, we still find recognizable variation in the distribution of frequencies of the phonemes. There are reasons expect this will be the case. Though phoneme inventories in Australia are similar, the phonemes themselves are known to exhibit considerable variation in their frequency distributions (Gasser and Bowern, 2014; Macklin-Cordes and Round, 2015). Likewise, phonemic bigram frequencies in the large Pama-Nyungan family exhibit diversity with a strong phylogenetic signal (Macklin-Cordes et al., accepted), suggesting that variations in Australian phonological frequencies have evolved over a deep time span. The main limitations of examining an Australian dataset is that Australian languages will lack many phoneme types (such as ejective stops and front rounded vowels) that may be found elsewhere in the world, and may have different characteristic frequencies of some individual phonemes (such as high frequency velar nasals, or low frequency long vowels); on the other hand, this is equally true of the Eurasian languages which have been the dominant basis of prior research. On balance, we find it worthwhile to increase the scope of knowledge to a new continent, and as the results will show, Australian phoneme inventories behave in ways that are clearly recognizable from Eurasia, suggesting that there are important, fundamental dynamics at play in the shaping of phoneme system that transcend the specifics of which phonemes are involved.

The phonemic frequencies in this study are extracted from wordlists. A consequence of this is that in our data, all lexemes are counted just once and so are weighted equally.\footnote{There is a significant body of research suggesting that frequencies defined in this manner are implicitly accessible to speakers and thus psychologically real (for example, Coleman and Pierrehumbert, 1997; Zuraw, 2000; Ernestus and Baayen, 2003; Albright and Hayes, 2003; Eddington, 2004; Hayes and Londe, 2006).} This differs from earlier work which has studied the frequencies of phonemes in discourse. In discourse, lexemes have unequal frequencies and so are weighted unequally. Indeed, since lexemes are known to have a Zipfian frequency distribution in discourse, one might wonder whether our lexical phoneme frequency data, which is not affected by these discourse effects, should correspondingly be expected to follow a different distribution -- perhaps a less Zipfian one. While the concern is understandable (we held it ourselves initially), the reality is that mathematically, the expected difference between phonemes' lexical and discourse frequencies \emph{on average} is zero (see Section S1, Supplementary Materials for a more extended discussion). So, although lexical and discourse frequencies will differ, their distributions will not differ systematically purely as a result of word frequencies being distributed in any certain way in discourse. Having said that, there could still be systematic differences due to other factors. For instance, if words with high discourse frequency were especially likely to contain the language's least-frequent phonemes, this could give rise to systematically different distributions; note however, that this effect would not be due merely to words having unequal frequencies in discourse, but be due to an additional factor, which links words' discourse frequencies directly and exceptionally to phoneme frequencies. While we cannot rule out the possibility of such additional effects, it remains true that the neutral expectation is that lexical and discourse phonemes frequency will be distributed similarly. Moreover, our study reveals results that are consistent with this neutral expectation. Our confirmation that this similarity in distributions is expected to be the case, and actually is the case, sets up a useful context for future studies also. For most languages of the world, discourse frequency data is not yet available, but wordlists are. As research into frequency distributions extends in scope to cover more of the world's languages, it is valuable to have clarified how results based on lexical frequencies and discourse frequencies relate to one another.

Our data comes from the Ausphon-Lexicon database, under development by the second author (Round, 2017b). Ausphon-Lexicon extends the Chirila resources for Australian languages (Bowern, 2016). It adds additional varieties and applies extensive data scrubbing, manual and automatic error-checking, and phonemic conversion using language-specific orthography profiles (Moran and Cysouw, 2018). A standing challenge for typological phonemic research is the long-recognized fact that phonemic analysis itself is non-deterministic (Chao, 1934; Hockett, 1963; Hyman, 2008; Dresher, 2009). Presented with identical sets of language data, two linguists may produce differing phonological analyses, not due to any error on the part of the linguist but due to differing applications of the multitude of criteria by which decisions are made during the analysis of a phonemic system. As a consequence, cross-linguistic phonological variation can be attributed not only to language facts, but also to variation in linguistic practice. In cross-linguistic research, it is desirable for information to be represented in a comparable way throughout a dataset, and so recent phonological literature has emphasized the value of \emph{normalizing} source descriptions prior to cross-linguistic analysis (Lass, 1984; Hyman, 2008; Hulst, 2017; Round, 2017a; Kiparsky, 2018). Phonemic representations in Ausphon-Lexicon are normalized in this sense. Section S3, Supplementary Materials, details the normalizations applied, together with bibliographic details of original data sources.

\begin{figure}

{\centering \includegraphics[width=180mm]{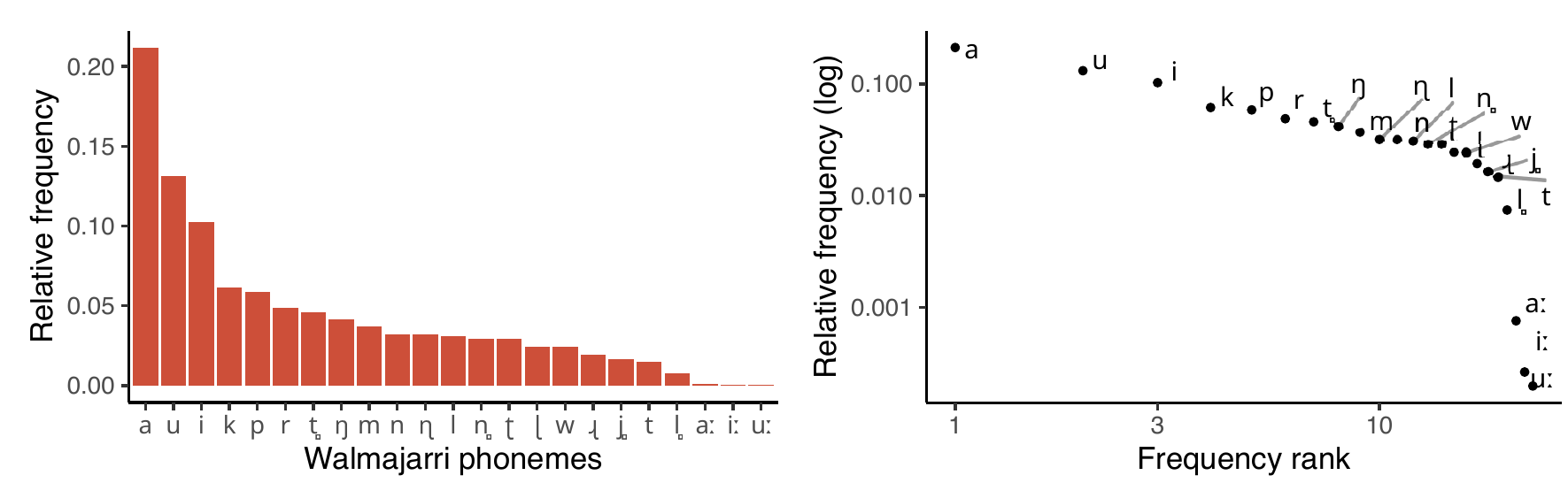} 

}

\caption{Frequency of phonemes in Walmajarri lexicon (Hudson and Richards, 1993). Plot (A) displays relative frequencies of each segment type. Plot (B) shows the same frequencies on log-transformed \(x\) and \(y\) axes---the traditional visual device used to identify power laws.}\label{fig:Figure-1}
\end{figure}

To illustrate an example of the phoneme frequencies in our sample, Figure \ref{fig:Figure-1} plots the frequencies of phonological segments in the Walmajarri lexicon (Hudson and Richards, 1993). Equivalent plots for every language in our sample can be viewed through an interactive visualization app that we provide in Section S6 of the Supplementary Materials.

Phonological frequency data differs in some respects from the data types most commonly encountered in scientific power law studies, such as word frequencies or city populations. Typically, in order to understand a population (and some property of it), such as the cities in the United States (and their sizes), or the words of English (and their frequencies), it is impractical to examine every last member of the population, and so the study will examine a sample. Ensuring that a sample is of sufficient size is an important consideration, firstly in order to adequately represent the population and additionally because a sufficiently large sample size is an important requirement in maximum likelihood estimation (Barndorff-Nielsen and Cox, 1994; Newman, 2005). In contrast, the phonemic inventory of any language is relatively small, and it is entirely feasible to examine exhaustive populations of phonemes.\footnote{The probability that we have failed to observe some phoneme that exists in a language is small, though non-zero. In Section S5, Supplementary Materials we evaluate whether this is the case in our data. We find only three languages where it is. Even in such cases, the missing segment inevitably will be an especially low frequency type, and is unlikely to dramatically alter the overall frequency distribution of segments in the language.} An advantage of this is that the sample is highly representative of the population, but a disadvantage is that the number of observations is small and cannot be increased.

Given a sample of phonemes, we require an estimate, or measurement, of their frequencies. Measurement error is a potential concern in this study. Our segment frequencies are calculated from documented wordlists, which necessarily are limited representations of the complete vocabulary of the languages that the wordlists represent. One concern is that the particular morphology of a language's citation forms may cause certain segments in the language to be overrepresented in a wordlist which contains only citation forms. This would represent a bias, that is, a factor that pushes observations in a certain direction. We have attempted to control for this, by removing identifiable citation-form tense morphology from verbal words and noun-class prefixes from nominals. Another source of concern is that wordlists with a smaller number of words will necessarily entail a greater level of uncertainty in the observed segment frequencies. This will be a source of noise in the data. It does not push observed frequencies in any particular direction, but makes them generally less accurate. To address this, in our study, we restrict the language sample to language varieties with a minimum wordlist size of 250 lexical items. We selected 250 lexical items as a cut-off on the basis of Dockum and Bowern (2019), who investigate the effect of wordlist size on phonological segment frequencies. Dockum and Bowern (2019) report accelerating losses in the fidelity of segment frequency estimates as a wordlist drops below 250 items. While more words will always yield better frequency estimates, we select a minimum of 250 as a reasonable compromise. This gives us a sample of 166 Australian language varieties. Wordlist sizes range from 268 to 8742 (median 1072, mean 1438). \newline

\hypertarget{method}{%
\subsection*{Statistical framework}\label{method}}
\addcontentsline{toc}{subsection}{Statistical framework}

We test for the presence or absence of a power law in the distributions of phonological segments following the maximum likelihood framework described by Clauset et al. (2009). In brief, Clauset et al.'s (2009, 663) proposed procedure consists of three steps:

\begin{enumerate}
\def\labelenumi{\arabic{enumi}.}
\item
  Estimate the parameters \(x_{min}\) and \(\alpha\) of the power law model using the maximum likelihood method (Barndorff-Nielsen and Cox, 1994; Newman, 2005)\footnote{Maximum likelihood estimation (MLE) is a method for estimating the parameters in a statistical model, given some set of observations by finding the set of parameter values, \(\hat{\theta}\), that maximize a likelihood function, \(P(x\ |\ \hat{\theta})\), where \(x\) is a set of observations. In our case, the parameters, \(\hat{\theta}\), to be estimated are those which define a particular distribution---for example, \(\alpha\) and (optionally) \(x_{min}\) in a power law model.}.
\item
  Calculate the goodness-of-fit between the data and the power law using the Kolmogorov--Smirnov (KS) statistic, where a larger value corresponds to a worse fit. Using a Monte Carlo procedure, a \emph{bootstrapped} \(p\) \emph{value} is calculated\footnote{This is a well-established statistical technique. A large number of simulated datasets are created, with data points drawn from the model power law distribution hypothesized in step 1. Each is then fitted to its own power law model and a KS statistic is calculated for the simulated dataset, relative to this model. The \(p\) value is defined as the fraction of these simulated KS distances larger than the actual, observed KS distance.}, and used to evaluate the plausibility of the power law. Namely, if this \(p\) value falls below a plausibility threshold of 0.1, the power law model is rejected.\footnote{Here we follow the method of Clauset, Shalizi, and Newman (2009), who suggest a threshold of 0.1. Note though, that even when p\textgreater{}0.1, we still do not necessarily accept that the power law is a good fit, rather there is a further round of evaluation (step 3). This use of a `\(p\) value' differs from the more common use case where a null hypothesis is rejected when the p value is above a certain level. The reason for the difference lies in how the hypothesis of interest is related to the null hypothesis. Commonly, the hypothesis of interest is set up as the alternative hypothesis, and low p-values are required to reject the null hypothesis (not of interest). Here, the hypothesis of interest (power law is plausible) is set up as the null hypothesis. Accordingly, it too is rejected when the p-value is low. By allowing it to be rejected all the way up to 0.1 (rather than 0.05, for example), we are setting the bar relatively high. This approach may seem counterintuitive in the context of testing a single distribution hypothesis (where it might seem better to make the distribution of interest deliberately harder to accept than to reject). But in the context of testing which distribution fits the data best among multiple alternatives, it makes sense to make it deliberately hard to reject any particular distribution type.} Otherwise, the power law model remains an initially plausible hypothesis, and we proceed to step 3.
\item
  Compare the power law model with a set of models representing alternative hypotheses. For each alternative model, a bootstrapped \(p\) value is calculated as in steps 1 and 2 above. A likelihood ratio test is performed, comparing the fit of the alternatives with those of the power law model. If the calculated likelihood ratio is significantly different from zero, this indicates a significant difference in plausibility, and its sign (positive or negative) indicates which model is favoured (Clauset et al., 2009, 680).
\end{enumerate}

We use the \emph{poweRlaw} package (Gillespie, 2014) in \emph{R} (R Core Team, 2017) to infer all maximum likelihood estimates and conduct bootstrapping to derive \(p\) values. We run 10,000 bootstrap iterations per language, per distribution type.\footnote{We find that 10,000 iterations are sufficient to obtain stable parameter estimates. Beyond 10,000 iterations, estimates will continue to fluctuate but in a tightly prescribed range. Plots of all bootstrapping runs can be viewed in the interactive visualization app provided in Section S6.}

\begin{figure}

{\centering \includegraphics[width=85mm]{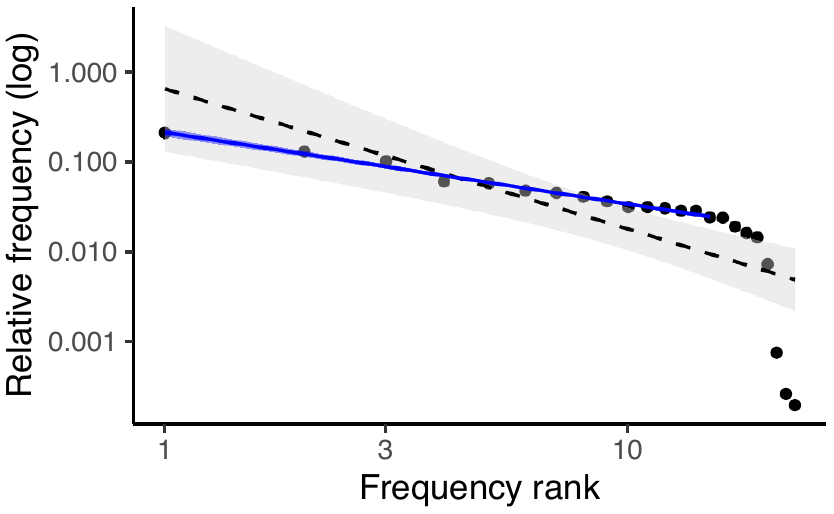} 

}

\caption{Log-log plot of frequencies versus frequency ranks in Walmajarri. When a linear model is fitted to the full distribution (dashed black), high and low frequency segments are overestimated and mid-rank segments are underestimated. When lowest-frequency segments are removed from the model (solid blue), the model appears to fit well.}\label{fig:Figure-2}
\end{figure}

As a brief point of comparison to prior work, we return to the Walmajarri example and plot the linear relationship between phoneme frequencies and rank on a log-log plot. Tambovtsev and Martindale (2007) find that a Zipfian distribution consistently underestimates the frequency of both high- and low-ranking segments while overestimating the frequency of those in the middle. The dashed black slope on Figure \ref{fig:Figure-2} shows a similar pattern. However, when the five lowest-frequency segments (i.e., those with the greatest statistical rank) are removed from the equation, the linear model fits much better (solid blue line). This is consistent with the observation by Clauset et al. (2009) that, in practice, power laws are rarely observed across the whole distribution---rather, there is a threshold, the \(x_{min}\) parameter, below which the power law ceases to apply. Visual inspection of other languages in the dataset indicates that Walmajarri's pattern of phoneme frequencies is common, although there is a good deal of variation (and, consequently, variation in the fit of a linear model). However, given the known limitations of applying a linear model to a log-log plot, we now turn to more reliable methods for validating the presence of a power law, using the maximum likelihood method outlined above.

\hypertarget{results}{%
\section*{Results}\label{results}}
\addcontentsline{toc}{section}{Results}

We firstly infer the fit of a power law to the full distribution of phoneme frequencies for each language, without estimating an \(x_{min}\) parameter. In Table \ref{tab:pl-summary} we summarize the maximum likelihood estimates of the power law distribution's defining shape parameter, \(\alpha\), the goodness-of-fit of the estimated power law distribution to the observed distribution of phoneme frequencies, and bootstrapped \(p\) values for the null hypothesis that the data are plausibly drawn from a power law distribution.

Mean \(\alpha\) is 1.38 (SD 0.17). As discussed earlier, the standard range of \(\alpha\) is 2 \textless{} \(\alpha\) \textless{} 3 (Clauset et al., 2009, 662). \(\alpha\) falls within this range for only 1 language. Furthermore, \(p\) values are very low. Just 2 of the 166 languages give a \(p\) value above the plausibility threshold.

Throughout this study, the possibility of type I error (false positives) must be taken into consideration. By setting our implausibility range at \(p \leq\) 0.1, we accept a one in ten chance of incorrectly rejecting a power law hypothesis which in fact is plausible---this can occur when the distribution's poor fit is due to chance fluctuation alone. Given 166 tests (one test per language), we would therefore expect to reject \(H_0\) incorrectly in around 17 (10\%) of those tests. In this instance though, we have rejected \(H_0\) as implausible in 99\% of the language sample. Thus it is clear that the power law distribution is being deemed implausible not merely by chance. It is genuinely a poor fit for the vast majority of languages. This result accords well with earlier work which has found that a simple, one-parameter form of the power law distribution poorly characterizes phoneme frequencies (Sigurd, 1968; Martindale et al., 1996; Tambovtsev and Martindale, 2007).

As discussed earlier, our dataset of phoneme frequencies is very likely to contain the complete population of phonemes in each language. At the same time, the number of observations per language is low---ranging from 16 to 37 segments in our language sample (mean 24.5, SD 3.8). Such a small set of observations can be a barrier to highly accurate maximum likelihood estimation. Clauset et al. (2009, 669) suggest that a minimum sample size of around 50 is needed to get a maximum likelihood estimate of \(\alpha\) accurate to at least 1\%. This is simply not possible for most of the world's languages (including all languages in this study) due to the limited size of segment inventories. Thus, in phonemic studies such as ours there is likely to be an unavoidable uncertainty in the estimate of \(\alpha\). \newline

\hypertarget{power-law-xmin-results}{%
\subsection*{Power law distribution with xmin}\label{power-law-xmin-results}}
\addcontentsline{toc}{subsection}{Power law distribution with xmin}

If the power law distribution, as inferred above, is inadequate for characterizing phoneme frequencies, then what other options are there? There are a couple of approaches to this question. One is to add an additional parameter to improve the fit of the power law; the other is to consider alternative distribution types. In this and the following sections we explore both approaches.

Here, we infer the fit of a power law distribution with an additional \(x_{min}\) parameter, whose effect is to remove some of the least-frequent observations from the sample which is being fitted. As above, we use maximum likelihood to infer the best-fitting \(x_{min}\) threshold for each language. Results are summarized in Table \ref{tab:pl-xmin-summary}.

After inferring an \(x_{min}\) parameter, the power law distribution is fitted to an average of only 13.9 segments, though there is a wide degree of variation (SD 4.1). In percentage terms, the power law distribution is fitted to an average of 57\% of a language's segmental inventory (SD 16\%). 117 languages (70\%) fall within the normal 2--3 range for \(\alpha\). Having only a small number of included observations above the \(x_{min}\) threshold can drive unreasonably high estimates of the \(\alpha\) scaling parameter. A sizeable portion of our sample (43 languages, 26\%) fall in this high range with \(\alpha\) above 3. At the other extreme, 6 languages (4\%) have an unusually low \(\alpha\) under 2. Mean \(\alpha\) is 2.75 (SD 0.65).

When \(x_{min}\) is included, the power law hypothesis is accepted as plausible (though, to emphasize, not necessarily correct) in the 158 of 166 language varieties for which \(p >\) 0.1. \(p\) falls below the 0.1 plausibility threshold in the remaining 8 languages. The lowest \(p\) value for any language is 0.011. This puts the chance of incorrectly rejecting \(H_0\) at around 1 in 100. The likelihood of a 1-in-100 event is high in a set of 166 tests. Overall, since the number of \(p\) values below 0.1 is considerably fewer than the number we would expect to observe through chance, and since there is a reasonable possibility that the lowest \(p\) value, 0.011, is a type I error, we cannot confidently rule out the power law hypothesis for any language in our sample.

Although we have failed to rule out the power law distribution as implausible for any specific language, this still does not mean that the power law distribution is the optimal one for our data, and there are some important caveats to our results so far.

A distribution will always fit a set of data at least as well as the same distribution with one fewer parameter. Thus, the observation that the power law distribution fits better when \(x_{min}\) is added requires some interpretation. Of greatest interest in this respect is the striking degree of improvement in fit, such that the power law distribution shifts from a largely implausible fit against full phoneme inventories, to a largely plausible fit after we exclude the least-frequent observations from samples. This raises the obvious question of why this might be so. We consider this in our Discussion, after we have also examined distributional alternatives to power laws.

The inclusion of an \(x_{min}\) parameter when fitting power laws is common practice, but its use is most obviously motivated in contexts where there are very many possible observations. For example, Clauset et al. (2009, 684) fit a power law to frequencies of unique words in Moby Dick and find a best-fitting \(x_{min}\) of 7 (\(\pm2\)). Words occurring fewer than 7 times can be disregarded and this still leaves nearly 3,000 unique words to which the power law distribution can be fitted. In contrast to this typical use case, where a large number of observations remain in play and do fit the power law, our use of \(x_{min}\) with phoneme datasets results in the exclusion of data points from an already small sample, leaving an even smaller set of data being fitted. As a general fact, it is inherently difficult to identify the most appropriate distribution for a small collection of observations. Correspondingly, it is not automatically an insightful finding, that a power law can be plausibly fitted to such small datasets. However, as mentioned just above, it is noteworthy that the same power law did not fit well to the slightly larger datasets that were being used without the \(x_{min}\) parameter. This suggests that it is not the merely small size of the dataset which is causing the good plausibility of the fit.

Small samples of observations can inflate \(p\) values, as is the case when investigating phonemes. We have good reason to suspect our \(p\) values are being inflated by the low number of observations per language, the evidence being that the number of \(p\) values we observe below 0.1 is considerably fewer than we would expect by chance. The difficulty we find in ruling out the power law distribution may reflect this. \newline

\hypertarget{alternative-distributions}{%
\subsection*{Alternative distributions}\label{alternative-distributions}}
\addcontentsline{toc}{subsection}{Alternative distributions}

\begin{figure}

{\centering \includegraphics[width=180mm]{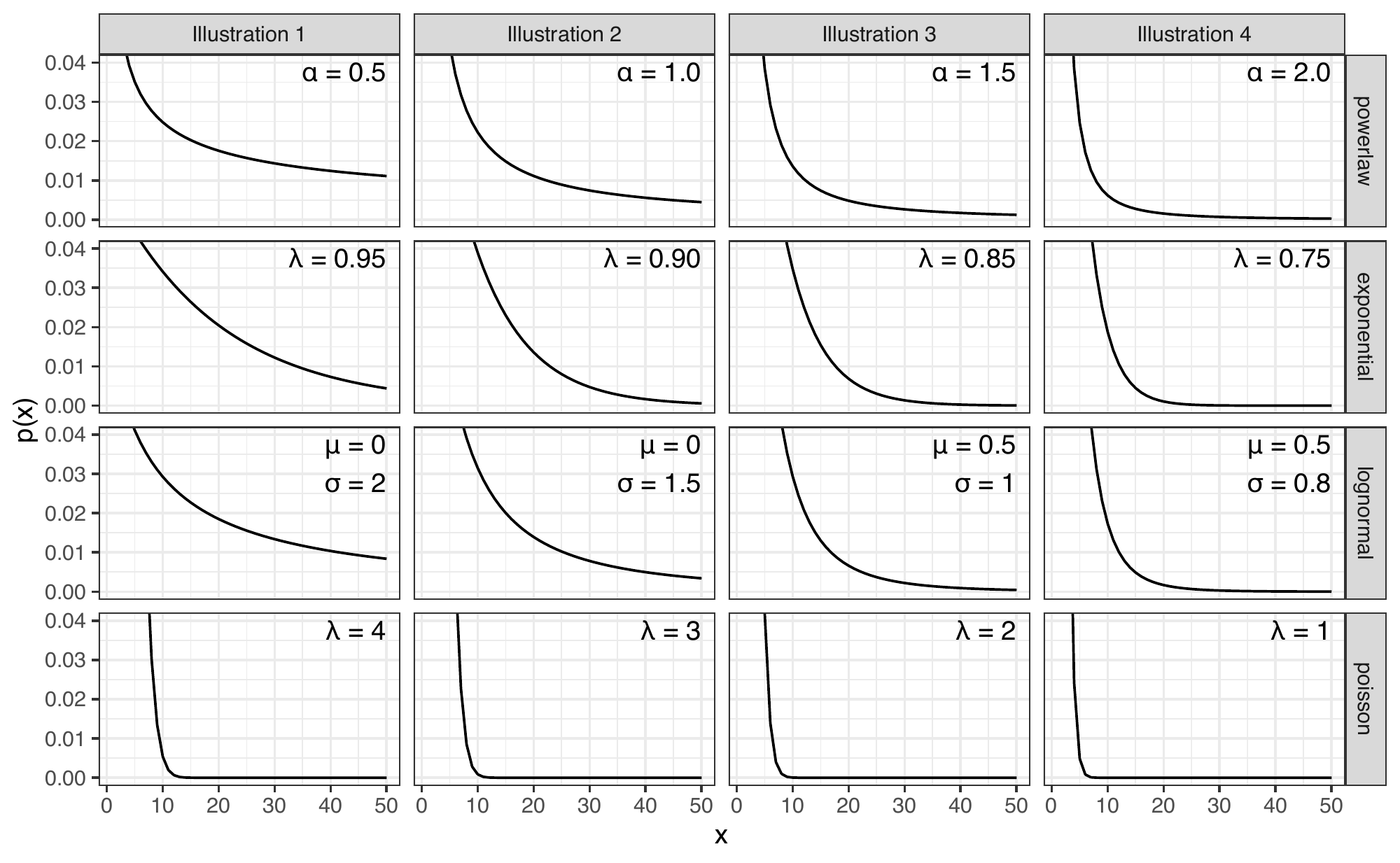} 

}

\caption{Four distribution types: power law, exponential, lognormal and Poisson, each illustrated with four parameterizations.}\label{fig:Figure-3}
\end{figure}

In addition to considering the merits of adding extra parameters to a distribution, we must also consider whether a completely different distribution would provide an equally good or better fit to the data. We consider three alternative distributions, which are not part of the power law family and may suggest different underlying generative processes. These are the lognormal, exponential and Poisson distributions. Like the power law distribution, the shape of these distributions can have a sharp initial peak and a rapidly decaying tail, as illustrated in Figure \ref{fig:Figure-3}. \newline

\hypertarget{lognormal-distribution}{%
\subsection*{Lognormal distribution}\label{lognormal-distribution}}
\addcontentsline{toc}{subsection}{Lognormal distribution}

The lognormal distribution is one where the data form a normal distribution when transformed on a log scale. Once again, we use the \emph{poweRlaw} package (Gillespie, 2014) to estimate parameter values using maximum likelihood. In this instance, the parameters to be estimated are log mean and log standard deviation parameters---the log-scale equivalent of the two parameters that define a normal distribution. We fit the distribution to the whole set of segment frequencies for each language---we do not estimate an \(x_{min}\) parameter at this stage (though see below). The lognormal distribution narrowly construed is a continuous distribution, however the \emph{poweRlaw} package contains a corresponding discretized version, appropriate to phoneme frequency data.

As for the power laws above, we calculate bootstrapped \(p\) values to assess the plausibility of the fit of the lognormal distribution for each language. The \(p\) values obtained are highly variable throughout the dataset. There are 73 languages (44\% of the language sample) for which \(p\) falls in the range of implausibility, below 0.1. This is higher than we would expect if the lognormal distribution were plausible for all languages and \(p \leq\) 0.1 values were due to type I error alone. This result is a little difficult to interpret, given the previously discussed difficulties with small samples of observations per language. What seems clear is that, given the rate of \(p \leq\) 0.1 values is elevated beyond chance, we cannot say that the lognormal distribution plausibly characterizes the segment frequencies of all languages. Nevertheless, for many languages---56\% of languages in our sample---we cannot confidently rule out the lognormal distribution. Overall, this makes the lognormal distribution with no \(x_{min}\) a better fit than the power law distribution with no \(x_{min}\), which we ruled out for up to 99\% of languages in the sample. One caveat to keep in mind is that the lognormal distribution is minimally defined by two parameters rather than one, which potentially puts it at an advantage compared to the single-parameter power law distribution. \newline

\hypertarget{exponential-distribution}{%
\subsection*{Exponential distribution}\label{exponential-distribution}}
\addcontentsline{toc}{subsection}{Exponential distribution}

An exponential distribution, and its discrete analogue, a geometric distribution, is one in which frequencies decay at a constant proportional rate. Thus for the frequencies of any two successively-ranked phonemes, \(x_k\) and \(x_{k+1}\), the distribution is characterized by a rate parameter, \(\lambda\), so that \(x_{k+1} = \lambda.x_k\). Here, as above, we use maximum likelihood to estimate the parameter \(\lambda\) and the bootstrapping procedure to obtain a \(p\) value.\footnote{This generates a relationship where \(p(x_k) \propto \lambda^k\). An alternative expression of the same relationship is \(p(x_k) \propto e^{-\lambda' k}\), where \(\lambda' = -log(\lambda)\). Results reported in Sections S5 and S6 of the Supplementary Material use this second definition, with \(\lambda'\).}

Bootstrapped \(p\) values are above the 0.1 plausibility threshold for 147 of 166 languages. The number of languages for which \(p \leq\) 0.1 is 19, close to the 17 or so that we would expect from type I errors. This, on the face of it, seems to make the exponential distribution quite a plausible model for phonological segment frequencies more generally. It must be noted, however, that there are a few languages for which the exponential distribution is a very poor fit. The most extreme, Miriwoong, has a goodness-of-fit statistic of 0.27 and a \(p\) value of 0.036. The poor quality of fit is visually evident on a log-log plot (see Section S6, Supplementary Materials). \newline

\hypertarget{poisson-distribution}{%
\subsection*{Poisson distribution}\label{poisson-distribution}}
\addcontentsline{toc}{subsection}{Poisson distribution}

The final distribution we consider is the Poisson distribution, which is related to the exponential distribution. The Poisson distribution is typically used to model the frequency of an event within some interval of time or space. Our case is a bit different since we are modelling the relationship between the frequency of many different events (different phonological segments) and their frequency rank in a language's phonological inventory. As with the exponential distribution, we use maximum likelihood to estimate a single parameter, \(\lambda\), and use bootstrapping to obtain a \(p\) value for the plausibility of the distribution.

The Poisson distribution is totally implausible for all languages in our language sample. Goodness-of-fit statistics range from 0.43 to 0.75 (mean 0.59, SD 0.07). We find \(p\) values indistinguishable from 0 in all cases. \newline

\hypertarget{summary-of-results-by-individual-distribution-type}{%
\subsection*{Summary of results by individual distribution type}\label{summary-of-results-by-individual-distribution-type}}
\addcontentsline{toc}{subsection}{Summary of results by individual distribution type}

In Table \ref{tab:results-summary}, we summarize results for the four distribution types evaluated in this study. For each distribution type, we give the number of languages for which the distribution's fit was deemed plausible (\(p >\) 0.1). For completeness, we give results for the exponential, lognormal and Poisson distributions when \(x_{min}\) is included, just as we did for the power law distribution. (Note: for one language, the bootstrapped \(p\) value estimation procedure failed to converge for the lognormal distribution with \(x_{min}\). This is the only distribution we tested which has three free parameters, and in this instance, the algorithmic procedure struggles to differentiate solutions with very similar likelihoods.) Perhaps most noteworthy among these results is the greatly increased inconclusiveness of the method when applied to the reduced set of data points lying above the \(x_{min}\) threshold. When the fitting task is restricted to a subset of only the most frequent segments in a language, it is possible to plausibly fit all but the Poisson distribution to any language, after type I error is factored in. One difference, which we nuance further in the next section, is that power law distributions with \(x_{min}\) are fitted on average to only 57\% of a language's phonemes, whereas the lognormal and exponential distributions are fitted to closer to 80\%. \newline

\hypertarget{evaluation-of-comparative-best-fit}{%
\subsection*{Evaluation of comparative best fit}\label{evaluation-of-comparative-best-fit}}
\addcontentsline{toc}{subsection}{Evaluation of comparative best fit}

The third and final step in Clauset \emph{et al.}'s (2009) framework is a likelihood ratio test. This third step may not always be necessary. If the bootstrapping procedure, above, were to show that only one distribution type plausibly fits the data, it would already have been shown to have the best fit among the candidates examined. However, bootstrapping may identify multiple distribution types as plausible. It will be recalled that just because a distribution is judged plausible via the bootstrapping process does not mean that it is the optimal one, since there may be other equally or more plausible distributions. Accordingly, when there are multiple plausible candidate distributions, Clauset et al. (2009) recommend using Vuong's (1989) likelihood ratio test for model selection, to determine the best-fitting of any pair of competing models. Full results of all likelihood ratio tests described in this section are tabled in Section S5 of the Supplementary Materials.

Vuong's (1989) test uses the Kullback-Leibler Information Criterion (Kullback and Leibler, 1951) to calculate the log likelihood of observing the data given a distribution model, and compares this to the log likelihood of observing the same data given a competing distribution model. The test returns a test statistic, which gives an indication of how strongly one model is favoured over another, and a \(p\) value, indicating whether the difference in the support for each model is statistically significant.

We begin by comparing distributions without the \(x_{min}\) parameter. As summarized in Table \ref{tab:results-summary}, two of these distributions (the power law and Poisson distributions, without \(x_{min}\)) have already been rejected as implausible for all or nearly all languages. Accordingly, we conduct just one likelihood ratio test per language, comparing the fit of the exponential versus lognormal distributions. Overall, we find that Vuong's likelihood ratio test somewhat favours the exponential distribution. Likelihood ratios favour the exponential distribution for 122 languages, and the lognormal distribution for 44 languages. However after Bonferroni correction, the difference in the likelihood of exponential and lognormal models is statistically significant for only two languages, Thaynakwithi and Dalabon, both favouring the exponential distribution.

Turning to distributions with the \(x_{min}\) parameter, since we have already rejected the Poisson distribution, we conduct likelihood ratio tests pairwise among the remaining three distributions. In order to compare distributions with \(x_{min}\) parameters, it is necessary to set \(x_{min}\) to the same value in both distributions (Gillespie, 2014). Thus, to make a pairwise comparison, we take the \(x_{min}\) value from distribution A and using it, re-estimate the other parameters of distribution B, and conduct one likelihood ratio test. Then we take \(x_{min}\) from B, use it and re-estimate the other parameters of distribution A, and conduct a second likelihood ratio test, giving two results for each pair of distributions.

Comparing the exponential and lognormal distributions, the likelihood ratios favour the lognormal distribution (139 languages to 27) using \(x_{min}\) from the lognormal fit, and favours the exponential distribution (103 languages to 63) using \(x_{min}\) from the exponential fit, however none of these comparisons reaches significance after Bonferroni correction.

Comparing the power law and lognormal distributions, likelihood ratios favour the lognormal distribution (144 languages to 22) using \(x_{min}\) from the power law fit, and all languages when using \(x_{min}\) from the lognormal fit, however only two of these comparisons reaches significance after Bonferroni correction. Yir Yoront favours the power law when using \(x_{min}\) from the power law fit and Malyangapa favours the lognormal distribution using \(x_{min}\) from the lognormal.

Comparing the power law and exponential distributions, the likelihood ratios favour the power law (133 languages to 33) when taking \(x_{min}\) from the power law fit, though no comparison reaches significance. They favour the exponential distribution 164 languages to 2 when \(x_{min}\) is taken from the exponential fit. Thirteen of those comparisons reach significance.

In sum, we found earlier that when parameterized without \(x_{min}\), only the exponential and lognormal distributions were broadly plausible. Voung's likelihood ratio test marginally favours the exponential test over the lognormal when fitted against entire phonemic inventories, but the difference is at most slight. When parameterized with \(x_{min}\), the power law distribution is fitted to around 60\% of languages' phonemes on average, while the exponential and lognormal are fitted to around 80\% (Table \ref{tab:results-summary}). Pairwise likelihood ratio tests, which apply one distribution's \(x_{min}\) parameter to the other, provide slender evidence of the following. Even when fitted against the small phonemic subsets favoured by the power law, the lognormal distribution may weakly outperform the power law, but the exponential distribution does not. Fitted against the larger subsets favoured by the exponential and lognormal distributions, the power law is outperformed by the exponential and lognormal. The performance of the latter two distributions is indistinguishable.

\hypertarget{discussion}{%
\section*{Discussion}\label{discussion}}
\addcontentsline{toc}{section}{Discussion}

Here we contextualize the current study more fully in the history of research into power laws and phoneme frequencies, before drawing implications and conclusions from the new findings we have obtained. \newline

\hypertarget{power-laws-linguistics}{%
\subsection*{Power laws in linguistics}\label{power-laws-linguistics}}
\addcontentsline{toc}{subsection}{Power laws in linguistics}

Investigation of power laws in the linguistic sphere has a long history. One of the oldest and best-known examples of a power law in any discipline is the distribution of word frequencies in text corpora, first noted by Estoup (1916) and subsequently described by Zipf (1932, 1949). Zipf's Law, as it has come to be known, is a discrete power law distribution. Its exponent parameter, \(\alpha\), is typically very close to 1, in which case, the second ranked item will be approximately half as frequent as the first, the third ranked item will be one third as frequent as the first, and so on. Zipf's Law continues to garner considerable attention, for example in Kucera et al. (1967), Montemurro (2001), and more recently in Baayen (2001, 2008). Various modifications to Zipf's formula have been suggested (notably Mandelbrot, 1954) and theoretical explanations put forward (Li, 1992; Naranan and Balasubrahmanyan, 1992, 1998).

Power laws have also been proposed to describe the distribution of phoneme frequencies. The use of Zipf's Law to model the frequencies of phonological segments initially appears to be an attractive prospect (Witten and Bell, 1990, 565--566). Nevertheless, a selection of alternative, non-power law distributions has also been suggested.

Sigurd (1968) is an early study evaluating the fit of a Zipfian distribution to phoneme frequencies, where the exponent, \(\alpha\), is set to 1. His evaluation method is a simple visual inspection, comparing observed phoneme frequencies in five languages (selected for their variety in segmental inventory size) with their expected frequencies assuming a Zipfian rank-frequency relationship. Sigurd (1968, 8) observes that the phoneme frequency distributions do not approximate a Zipfian curve, particularly for the most common segments. Rather, Sigurd (1968) finds better approximations using a geometric series equation, so that \(x_{k+1} = \lambda.x_k\) for some rate parameter \(\lambda\), giving the discrete distribution:

\begin{equation}
p(x_k) \propto \lambda^k
\label{eq:sigurds-geometric}
\end{equation}

Good (1969, 577) suggests an alternative method of approximation: following Whitworth (1901), Good calculates the expected frequencies of each phoneme given a process whereby a unit interval probability space \([0,1]\) is divided into \(n\) parts at random (where \(n\) is the number of phonemes in the language), following a uniform distribution. This is equivalent to a so-called stick-breaking process: imagine a stick, which represents the unit interval probability space. The stick is broken into \(n\) parts; the \(n-1\) places along the stick at which a break is made are selected randomly and all at once, with any place along the stick equally likely to be selected as any other. When these parts are rearranged by size, from smallest to largest, their expectation follows the distribution:

\begin{equation}
\frac{1}{n^2},\quad \frac{1}{n} \left( \frac{1}{n} + \frac{1}{n-1} \right),\quad \frac{1}{n} \left( \frac{1}{n} + \frac{1}{n-1} + \frac{1}{n-2} \right),\quad ...
\label{eq:whitworth-dist}
\end{equation}

Which is to say:

\begin{equation}
p(x_k) \propto \sum_{i=k}^{n} \frac{1}{i}, \quad 1 \leq k \leq n
\label{eq:whitworth-rank-dist}
\end{equation}

In support of this model, Good (1969, 577) provides a table of observed versus expected frequencies of both graphemes and phonemes in English, however the sample size is modest (1000 words) and does not extend to any other languages. Furthermore, there is no visual or statistical evaluation of the goodness-of-fit. Good (1969) intends for the results to be taken as a curious observation only, with no strong theoretical position or claim of generalizability.

As \(n\) in equation \eqref{eq:whitworth-rank-dist} grows large, the summation term \(\sum_{i=k}^{n} \frac{1}{i}\) converges towards \(-log(\frac{k}{n+1})\) (Loeb and Rota, 1989, 12), meaning that Good's distribution can be considered an approximation of \eqref{eq:borodovsky-rank-dist}, a discretized negative logarithmic distribution.

\begin{equation}
p(x_k) \propto -\log \frac{k}{n+1}, \quad 1 \leq k \leq n
\label{eq:borodovsky-rank-dist}
\end{equation}

Gusein-Zade (1988) and Borodovsky and Gusein-Zade (1989) visually evaluate the fit of \eqref{eq:borodovsky-rank-dist} to the graphemes of English, Estonian, Russian and Spanish.\footnote{Of course, the statistics of graphemes are different from the statistics of phonological segments. As Bloomfield (1935, 136--137) rather emphatically points out: ``If we take a large body of speech, we can count out the relative frequencies of phonemes and of combinations of phonemes. This task has been neglected by linguists and very imperfectly performed by amateurs, who confuse phonemes with printed letters.'' Nevertheless, the frequencies of graphemes has been of interest historically in many applications; for example, in traditional printing, the development of Morse code, and library cataloguing (Witten and Bell, 1990, 550--551).} They also use the equation to describe the distribution of DNA codons (Borodovsky and Gusein-Zade, 1989). Witten and Bell (1990, 563--566) examine the frequencies of single graphemes, graphemic bigrams and trigrams in the Brown Corpus and compare the fits of Good's distribution and Zipf's Law by comparing expected entropy values for each model to observed entropy scores. They find that the quality of the fit of Good's model declines with bigrams and trigrams compared to single graphemes, although the observed distribution curves are broadly of the same shape (and resemble the shape of Good's distribution rather than the Zipfian distribution). When assessed using metrics based on entropy, Good's distribution fits better than or around equally as well as the Zipfian distribution for all three datasets. Good's distribution and the negative logarithmic distribution it approximates also have the advantage of parsimony, since they are parameter-free: knowing how many unique items (phonemes, graphemes, bigrams, etc.), \(n\), are in the dataset is sufficient to calculate their expected distribution of frequencies---there are no additional parameters to estimate such as \(\alpha\) in \eqref{eq:zipfs-law} or \(\lambda\) in \eqref{eq:sigurds-geometric}.

Martindale et al. (1996) compare the fit of four different distributions to frequencies of both graphemes and phonemes in text corpora from 18 languages. Using the \(R^2\) statistic in a linear regression, they compare the fit of the parameter-free negative logarithmic equation of Borodovsky and Gusein-Zade (1989) in \eqref{eq:borodovsky-rank-dist} to the Zipfian power-law distribution \eqref{eq:zipfs-law}, Sigurd's geometric series distribution \eqref{eq:sigurds-geometric}, and the Yule-Simon distribution (Yule, 1925; Simon, 1955), which can be written:

\begin{equation}
p(x_k) \propto \frac{1}{k^\alpha}.\lambda^k
\label{eq:yule}
\end{equation}

The Yule-Simon equation in \eqref{eq:yule} is the product of the power law in \eqref{eq:zipfs-law} and the geometric equation in \eqref{eq:sigurds-geometric}. Because of the differing rates at which the two parts of the equation decay as \(k\) increases, equation \eqref{eq:yule} produces a distribution which is more like a power law \eqref{eq:zipfs-law} for low values of \(k\) (and thus for high frequency items, for instance) and more like the geometric \eqref{eq:sigurds-geometric} for high values of \(k\) (low frequency items) (Simon, 1955).

The Yule-Simon equation in \eqref{eq:yule} has not just one free parameter but two, the exponent \(\alpha\) and the rate \(\lambda\), and the Zipfian and Sigurd equations are effectively special cases of it, each with one parameter fewer. The Zipfian distribution is equivalent to \eqref{eq:yule} with \(\lambda\) set to 1 (so that \(\lambda^k = 1\)), while the geometric equation is equivalent to \eqref{eq:yule} with \(\alpha\) set to zero (so that \(1/k^\alpha = 1\)). This is important, since as a general fact, if distribution A is a special case of distribution B, with fewer free parameters than it, then B will always perform at least as well as A when fitting the same set of data. Thus, the Yule-Simon distribution will necessarily fit the same set of data at least as well as the Zipfian distribution, and Sigurd's geometric distribution.

Martindale et al. (1996) find that the Yule-Simon distribution fits best, for both graphemes and phonemes. They find that the Zipfian distribution tends to overestimate both high and low frequency items, although the differences they observe between models are only small. On this basis, they conclude that it is ``a matter of taste'' whether one opts for the more precise Yule-Simon distribution or simpler models with fewer parameters to estimate (Martindale et al., 1996, 111). Tambovtsev and Martindale (2007) expand Martindale et al.'s (1996) study to include phoneme frequencies in 95 languages (90 of these are Eurasian; 2 are from Oceania and 1 each from Australia, Africa and South America). The sample is divided into four language groups (Indo-European, Altaic and Yukaghir-Uralic--plus a miscellaneous group) and a series of pairwise sign tests are conducted to test whether the difference in mean \(R^2\) is significant between different distributions for each language group. Again, they find that the Yule-Simon distribution fits best overall.\footnote{Although in their statistical tests they do not adjust their significance levels to correct for multiple hypothesis testing.}

Obtaining a better fit by using a distribution with an additional parameter may be relatively trivial mathematically speaking, but this does not mean it is uninteresting. The extra parameter may work to capture a significant real-world nuance in an underlying causal process or describe the effect of one or more secondary processes. A compelling causal explanation of a complex distribution might therefore be formulated by identifying some real-world factor and explaining how its mathematical effect on the distribution is expected to match what we find. It is also important to consider the possibility of equifinality---the fact that multiple, different real-world phenomena may have equivalent mathematical effects. Tests of goodness-of-fit examine only the mathematical aspect, and cannot distinguish between different phenomena whose detectable mathematical contribution is equivalent.

Martindale et al. (1996, 111) and Tambovtsev and Martindale (2007, 9) note that a Zipfian distribution describes frequencies of phonological segments less well than it describes frequencies of words, in part because the highest-frequency phonemes are not frequent enough. They speculate that this may be so, because if the most-frequent phonemes did pattern in a Zipfian way, then perception problems could arise for language users owing to the small size of a phonological inventory. This speculation does not meet the criteria for a compelling causal explanation though. It is not clarified what the linguistic mechanism is, that acts to prevent such perceptual problems, and thus we do not have a real-world phenomenon whose mathematical properties could be interrogated. Nor is it explained why, if such a mechanism exists, its mathematical effect would be to contribute something like the extra geometric term \(\lambda^k\) that differentiates the Yule-Simon distribution \eqref{eq:yule} from the Zipfian \eqref{eq:zipfs-law}.\footnote{The Yule-Simon equation, which Martindale et al. (1996) and Tambovtsev and Martindale (2007) find to be a superior fit, describes a distribution which is most similar to a power law for high frequency (low \(k\)) items, and most like the geometric for low frequency (high \(k\)). The claim that its superior fit is due to \emph{non}-power-law-like behaviour of high frequency items is therefore hard to reconcile with the mathematics.} This is not to say that an explanation in terms of perceptibility and confusability is implausible, but rather if our aim is for causal hypotheses that can be evaluated, then more steps are needed, a topic that we return to below. Next, however, we consider the history of investigation presented here, together with our new results. \newline

\hypertarget{findings-from-a-more-reliable-evaluation-procedure}{%
\subsection*{Findings from a more reliable evaluation procedure}\label{findings-from-a-more-reliable-evaluation-procedure}}
\addcontentsline{toc}{subsection}{Findings from a more reliable evaluation procedure}

Power laws have attracted wide and sustained scientific interest. Recent debates on their validity have prompted the development and widespread adoption of novel evaluation methods that are more reliable than those used in the past (Clauset et al., 2009; Stumpf and Porter, 2012). In this study, we re-evaluated the plausibility of several distribution types as characterizations of phoneme frequencies using a maximum likelihood statistical framework presented by Clauset et al. (2009) and a sample of 166 Australian language varieties.

Using a more reliable evaluation procedure than previous investigations, we have confirmed the finding that a basic power law distribution, with a single free parameter, is generally insufficient for characterizing phoneme frequencies. Additionally, we reconfirm a result going back to Sigurd (1968), that an exponential (or geometric) distribution, with a single free parameter, is a good plausible fit for full phonemic inventories. Furthermore, we find that a lognormal distribution, with two free parameters, is an additional plausible fit, whereas a Poisson distribution, with a single free parameter, is implausible. We did not attempt to fit the two-parameter Yule-Simon distribution in \eqref{eq:yule}, since to our knowledge, there is no maximum likelihood estimation procedure currently available for estimating its parameters. However, we do return to the question of this distribution just below.

A second novel contribution was to consider the addition of an \(x_{min}\) parameter, a practice which is now common in power law research. Notably, while power laws are largely implausible fits for entire phoneme inventories, their plausibility is improved strikingly once a subset of the least-frequent phonemes is removed from the sample. This is despite that fact that the full inventories and the reduced ones share the property of comprising notably small samples. The subset removed in order to achieve maximum likelihood is on average large, at 43\%. This result indicates that power laws constitute a plausible characterization for the more-frequent portion of phonemic inventories, and explains why the upper end of a Yule-Simon distribution, which most closely approximates a power law, should be a reasonable fit. We note however, that the lognormal distribution also performs well in this same, high frequency region of phonemic inventories. Exponential (or geometric) distributions do not fit the higher-frequency portion of inventories as well the power law or lognormal do, but they are good fits for entire inventories, suggesting that they fit particularly well in lower-frequency portions. This would explain why the lower end of a Yule-Simon distribution, which most closely approximates a geometric distribution, should be a reasonable fit.

Using an evaluation procedure which has since been shown to be unreliable, Martindale et al. (1996) and Tambovtsev and Martindale (2007) concluded that the two-parameter Yule-Simon distribution fit whole inventories better than a power law or a geometric distribution. Implicitly, this is a conclusion in two parts: more-frequent phonemes are more power-law-like, and less-frequent are more geometric-like. Here we have not been able to directly evaluate the Yule-Simon distribution using the more reliable, maximum likelihood method. However, we have found evidence supporting a similar conclusion, that the more-frequent and less-frequent portions of phonemic inventories are characterized by different distributional properties. The more-frequent portion better matches a power law, though also a lognormal distribution. The less-frequent portion better matches a geometric distribution. These two findings serve to clarify and qualify the two implicit halves of the main finding of Martindale et al. (1996) and Tambovtsev and Martindale (2007), and here we have arrived at them by more reliable methods. Furthermore, by estimating \(x_{min}\) parameters, we have provided some estimates of where power-law-like behaviour starts to cut out within a phonemic inventory. To understand what these results entail for theory, we return to the question of causal processes. \newline

\hypertarget{distributions-outcomes-of-stochastic-processes-linked-to-causal-factors}{%
\subsection*{Distributions: outcomes of stochastic processes linked to causal factors}\label{distributions-outcomes-of-stochastic-processes-linked-to-causal-factors}}
\addcontentsline{toc}{subsection}{Distributions: outcomes of stochastic processes linked to causal factors}

Ultimately, linguistic theory seeks to explain the patterns that can be observed in human language. This endeavour will be aided by a sound knowledge of which mathematical distributions plausibly characterize a given variable \(x\) (such as phoneme frequency), since those distributions will be consistent with only certain mathematical kinds of underlying processes. In this paper, we have improved the certainty of our understanding of observed distributions of phoneme frequencies, using state-of-the-art statistical methods. This is a necessary step, but a first step only. A fruitful next step used widely in other sciences is to explicitly consider mathematical families of stochastic processes, whose signature outcome distributions are consistent with those of our empirical observations, and then to ask in turn, what plausible, real-world causal processes could be consistent with these observation-matching stochastic processes.

It can be emphasized that in this model of progress, a good understanding of families of stochastic processes will play an important, enabling role. Accordingly, it will be useful to harness advances that have already been made elsewhere. For instance, many discrete systems can be profitably conceptualized in terms of urn processes with characteristically associated distributions. As Kuba and Panholzer (2012, 87) remark, ``{[}u{]}rn models are simple, useful mathematical tools for describing many evolutionary processes in diverse fields of application''. There exist well-studied urn processes which yield many kinds of distributions. It will be profitable in linguistic research to more clearly relate our own theories of change, including change in phonemic inventories, to these mathematically more generalized processes. By doing so, linguists will be able to tap into related, existing mathematical results (such as relating processes to distributions), that can assist us to further differentiate the theories that are more viable from those that are less so. \newline

\hypertarget{casual-processes-of-phoneme-addition-removal-and-redistribution}{%
\subsection*{Casual processes of phoneme addition, removal and redistribution}\label{casual-processes-of-phoneme-addition-removal-and-redistribution}}
\addcontentsline{toc}{subsection}{Casual processes of phoneme addition, removal and redistribution}

Naturally, any observation-matching families of stochastic processes will still need to be related to linguistically plausible causal processes (Cysouw, 2009). Though our paper has primarily focused on methods that will strengthen future discussions around links between observations and causal mechanisms, rather than the causal mechanisms themselves, here we offer some brief remarks on historical mechanisms affecting phoneme frequencies. As we do, we continue using the imagery of urns to represent phonemes types; the balls in them to represent their tokens in the lexicon; and processes that serve to add, remove and redistribute them. (For a like-minded review of potential mechanisms behind Zipf's law in the frequencies of \emph{words}, see Piantadosi, 2014.)

Phoneme frequencies undergo constant modification due to changes occurring at multiple levels of linguistic structure, including the phonology proper, morphology, syntax and the grammar of discourse. Within the phonology itself, fundamental changes include deletions, insertions and changes of phonemic category, all of which will impact phoneme frequency distributions. In an urn model, historical deletions will result in balls being removed from a phoneme's corresponding urn while insertions will add balls. Adding a level of complexity, many phonological changes affect not just one phoneme category, but natural classes of sounds, thus deletions and insertions that apply to natural classes will remove or add balls simultaneously from a non-random, `natural' set of urns.

Phonemic mergers (Hoenigswald, 1965) are changes which collapse two erstwhile phoneme categories into one, effectively emptying all balls from one urn into another. Phonemic partial mergers, also known as primary splits, shift only some of a phoneme's instances into another, existing phonemic category, transferring some portion of an urn's balls to another. Phonemic (secondary) splits involve an existing phonemic category splitting into multiple new categories, entailing the creation of one or more new urns, filled with some proportion of the balls from an existing urn, which itself remains non-empty. In both mergers and splits, natural classes may be involved, entailing simultaneous transfers between, and creation or loss of, `natural' sets of urns.

A phonemic category will map onto multiple actual speech sounds, or phones, and the phonemic categorization of a phone frequently depends on its contextual environment. Consequently, it is not uncommon for multiple types of phonemic changes to occur at once, owing to the fact that when one sound changes, so too does the contextual environment of its neighbours. Under these circumstances, the ubiquity of non-uniform distributions of sounds in various contexts will lead to non-accidental, correlational relationships among these coupled changes.

At the morphological level, changes in the frequency of certain formatives will cause concerted frequency changes in the set of phonemes comprising the formative, though in this case, there is no expectation that the set involved will form a natural class. If the object of study is the discourse frequencies of phonemes, then similar effects will arise when the usage frequencies of words undergo change. Furthermore, lexicons are not closed systems. Words can be borrowed from other languages. The effects of borrowing on phoneme frequencies in the recipient language is a complex matter (Boretzky, 1991). Here we name just a few factors. The donor language will have its own phonemic repertoire and associated frequencies, which will bias what can be donated and at what relative rates. Phonemic borrowing is not direct, but is mediated by phonetic similarities and psychological equivalences drawn by speakers across languages (Flege, 1987; Kang, 2003). Correspondences between donor and recipient phonemes will therefore exhibit correspondences that are broadly natural (Paradis and LaCharité, 1997), yet the process is frequently variable (Lev-Ari et al., 2014) and may involve centuries of subsequent remodeling of borrowed material (Crawford, 2009). Moreover, additional systematic mutations, deletions and insertions of phonemes may be motivated by constraints on the permissible phonotactic (contextual) arrangement of phonemes in the recipient language.

There are implications to be drawn from this, for the explanation of distributions of phoneme frequencies. We turn to these implications below. \newline

\hypertarget{implications-for-explanatory-accounts}{%
\subsection*{Implications for explanatory accounts}\label{implications-for-explanatory-accounts}}
\addcontentsline{toc}{subsection}{Implications for explanatory accounts}

In most of the historical changes noted above, it matters not only whether, and how many, balls are moving from one urn to another, but exactly which real-world phonemes the urns themselves represent. Phonemes form natural classes, which may change in tandem. Phonotactic arrangements, which, notwithstanding variation, exhibit strong similarities cross-linguistically, will impact how changes mediated by contexts are correlated. And borrowing too operates in reference to phonemes' actual substance. This importance of phonemic substance raises a host of questions, which can be addressed in new and potentially revealing ways within a stochastic-modelling research program. For instance, the frequencies with which various kinds of phonemes are present or absent across the world's languages varies greatly (Maddieson, 1984; Moran, 2012; Everett, 2018a): what is the range of assumptions under which this result would emerge, within a stochastic model? Does it demand models with strong effects corresponding to well-known articulatory, acoustic and perceptual factors that cross-linguistically favour certain phoneme types (Liljencrants and Lindblom, 1972; Stevens, 1989; Johnson et al., 1993; Browman and Goldstein, 1986; Proctor, 2009; Becker-Kristal, 2010; Everett, 2018b) or can it be derived from weak ones? And can some effects arise entirely through indirect interactions? For instance, vowels typically stand adjacent only to consonants, while in most languages consonants may be adjacent to vowels or other consonants; can it be shown that this basic phonotactic asymmetry predicts differing rates of change among vowels and consonants (Moran and Verkerk, 2018)? Returning to the main topic of this paper, might we predict that languages can exhibit significantly different overall phoneme frequency distributions depending on which phonemes they contain? For example, might observations for one part of the world (such as the Australian data in this paper) therefore differ in an explicable manner from observations in others?

Even if we abstract away from issues related to phoneme substance, a successful stochastic model of phoneme frequency evolution may still require a mixture of sub-models, obeying different principles, given the existence of the multiple types of historical processes that affect phoneme frequency: deletion, insertion, splits, mergers, borrowings. Though it might be clear in outline what (some of) these sub-models must be, it is far less clear, empirically speaking, what their quantitative parameterizations are. In what precise mix do these sub-processes occur? How often do events occur independently or in concert? What empirical grounds do we have for estimating these parameters, and what are our levels of uncertainty? In these respects, there exists another interesting role to be played by stochastic modelling studies. We currently lack empirical answers to many of these quantitative questions, but can we reveal limits to what is plausibly compatible with observed phoneme distributions? A related open question, connected to issues raised in the previous paragraph, is to what extent a successful stochastic model will need to reflect empirical complexities, like phonemic substance, natural classes, phonotactics and concerted changes, as points of non-independence within and between sub-models, or can some of these be ignored with little impact on the outcomes predicted by the model? Whatever the answers may be, we see a productive and interesting field of inquiring lying directly ahead.

We conclude this section by returning to our empirical findings and situating them within the causal reasoning just outlined. Our empirical finding was that the more frequent phonemes in a language's inventory tend to be distributed more in line with a power law, and the less frequent more in line with an exponential distribution. As we mentioned earlier, power law distributions can be stochastically generated by preferential attachment processes. In addition to this, exponential distributions can be stochastically generated by a so-called birth-death process, in which entities arise at some characteristic birth rate \(\lambda\), and disappear at a characteristic death rate \(\mu\). It may be---and here we are speculating, but in a reasoned manner---that as tokens of phones shift between phonemic categories during the kinds of sound change processes we mentioned above, different phonemic categories have different probabilistic propensities to be source categories (i.e., the erstwhile category of the changing phone) and destination categories (its new category after the change). This alone would constitute a birth-death regime (Cysouw, 2009), and its causal roots would lie in phonemic substance. Taking this further, it may also be that there is a separate propensity for phones to migrate towards numerically stronger categories. This would contribute a preferential attachment dynamic, whose causal roots would be something over and above mere phonemic substance. An interesting point to note, is that under this scenario, it is not important which specific phonemes an inventory contains: so long as phonemes have differing propensities to be source or destination categories during changes, and so long as numerically stronger categories are favoured as destinations, the same outcomes should be obtained. This would explain how our Australian results can resemble earlier, Eurasian, results so closely. However, what remains incomplete in this picture is firstly: When a preferential attachment process and a birth-death process are simultaneously active in one and the same system, under what parameterizations of the system does it fall out that preferential attachment affects the more-frequent categories more strongly than the less-frequent, in a fashion similar to what appears to be the case in phoneme inventories? And secondly: What factor(s) might underlie both the preferential attachment dynamic itself and the overall parameterization of the system? There has been some work on this question, suggesting that climactic factors (namely the effect of humidity on vocal fold physiology) (Everett et al., 2015), universal sound-meaning associations (Blasi et al., 2016) and/or relative physiological ease of articulation (Everett, 2018a, 2018b) might drive a preferential attachment dynamic, in which languages gravitate towards certain phonemic categories on the basis of these various factors. It may be that different hypotheses about possible causal mechanisms for the preferential attachment dynamic entail different predictions on these crucial points. These ideas are worth chasing up further. \newline

\hypertarget{reasons-to-seek-wider-horizons}{%
\subsection*{Reasons to seek wider horizons}\label{reasons-to-seek-wider-horizons}}
\addcontentsline{toc}{subsection}{Reasons to seek wider horizons}

There are likely to be additional, valuable variables, and possibly more tractable variables, to study beyond just phoneme frequency distributions. In this paper we have focused on phoneme frequencies because they have occupied such a prominent place in the history of investigations of distributions. However, by doing so we do not wish to suggest that phonemes ought to continue to occupy such a prominent place. Firstly, we agree with Piantadosi (2014), that the focus of investigation should not be on distributions for their own sake, but on what a better understanding of distributions can tell us about language. It may be, that further study reveals that phoneme frequency distributions are not very powerful, when it comes to discriminating plausible from implausible causal models, whether this is because phoneme inventories are too small to allow their distributions to be characterized with sufficient precision or because those distributions themselves are consistent with too many competing explanations. However, the very same research may reveal other variables that are more valuable. To speculate, perhaps it will prove more interesting to ask how \emph{sets} of phonemes, such as natural classes, pattern within their languages' frequency ranks across languages; or to examine not merely phonemes (of which there are only few categories per language), but phonemes within certain contextual environments (of which there are more). Moreover, some interesting results may emerge only within models and statistical analyses that grant a role to phylogeny (Macklin-Cordes et al., accepted). \newline

\hypertarget{conclusions}{%
\subsection*{Conclusions}\label{conclusions}}
\addcontentsline{toc}{subsection}{Conclusions}

There are many branching paths of research ahead. Nevertheless, there will be some basic principles that we see remaining constant, and these have been the core focus of our contribution here. In this paper we have demonstrated and outlined a template for future work on distributions. Ideally, such work should begin with critical assessment of links that can be made between existing or new causal hypotheses, including diachronic processes, and from these, via stochastic models, to particular distributional outcomes. Subsequently, the fit of the hypothesized distributions to real-world data should be evaluated rigorously using robust statistical methods. Lastly, an attempt must be made to rule out competing distribution types and alternative generative mechanisms. As our investigation demonstrates, this may be challenging, given the inherent limitations of working with small sets of observations. Maintaining a clear-eyed view of these limitations, and using advances already made in allied fields, will help spur this field of inquiry to new, robust insights into the dynamics of phoneme inventories.

\hypertarget{disclosureconflict-of-interest-statement}{%
\section*{Disclosure/Conflict-of-Interest Statement}\label{disclosureconflict-of-interest-statement}}
\addcontentsline{toc}{section}{Disclosure/Conflict-of-Interest Statement}

The authors declare that the research was conducted in the absence of any
commercial or financial relationships that could be construed as a potential
conflict of interest.

\hypertarget{author-contributions}{%
\section*{Author Contributions}\label{author-contributions}}
\addcontentsline{toc}{section}{Author Contributions}

JM conceived and designed the study with guidance from ER. ER compiled the study's data. Together, JM and ER performed the analysis and interpreted the results. JM composed an initial draft of the manuscript, compiled the supplementary materials and produced the figures. ER critically revised the manuscript and supplementary materials, and contributed important intellectual content. Both authors approved the final version of the submitted manuscript and agree to be accountable for all aspects of the work.

\hypertarget{acknowledgments}{%
\section*{Acknowledgments}\label{acknowledgments}}
\addcontentsline{toc}{section}{Acknowledgments}

We wish to thank Caleb Everett and Steven Moran for their insightful comments on earlier drafts of the manuscript. Their feedback and guidance helped us to improve the paper considerably, and we appreciate the generosity of their time and effort.

Funding: JM is supported by an Australian Government Research Training Program Scholarship. Data compilation was funded by Australian Research Council grant DE150101024 to ER.

\hypertarget{supplemental-data}{%
\section*{Supplemental Data}\label{supplemental-data}}
\addcontentsline{toc}{section}{Supplemental Data}

Supplementary information and materials for this paper are organized into 7 parts.

Text-based supplementary information is contained in the PDF document \emph{S1-S5.pdf}, accompanying this paper. This document includes a discussion of neutral expectations about phonemes' lexical and discourse frequencies (S1), a guide to data and code (S2), a bibliographical list of original wordlist sources (S3), a comparison of phoneme inventories in Ausphon and Phoible 2.0 (S4), and full tables of results (S5).

Further supplementary materials are contained in a zip folder \emph{S6-S7\_suppmaterials}, hosted on Zenodo at \url{https://doi.org/10.5281/zenodo.4104116}. The \emph{S6\_data\_viewer} directory contains an interactive data viewer. The interactive data viewer is also hosted in an accompanying R package on Github (\url{https://github.com/JaydenM-C/phonfreq}) for ease of installation and use (see the description in S2 for further details). The \emph{S7\_data\_code\_results} directory contains 3 subdirectories. The \emph{data} subdirectory contains a tab-delimited format spreadsheet of phoneme frequency data. The \emph{R} subdirectory contains code used to perform analysis and create figures. The \emph{results} subdirectory contains the output of analysis saved as \emph{.Rdata} format objects for use with R statistical software.

\hypertarget{references}{%
\section*{References}\label{references}}
\addcontentsline{toc}{section}{References}

\medskip

\begingroup
\setlength{\parindent}{-0.2in}
\setlength{\leftskip}{0.2in}
\setlength{\parskip}{0pt plus1pt}

\noindent

\medskip

\begingroup
\setlength{\parindent}{-0.2in}
\setlength{\leftskip}{0.2in}
\setlength{\parskip}{0pt plus1pt}

\noindent

\hypertarget{refs}{}
\leavevmode\hypertarget{ref-albert_species_2011}{}%
Albert, J. S., Bart, H. J., Jr., and Reis, R. E. (2011). ``Species richness and cladal diversity,'' in \emph{Historical biogeography of neotropical freshwater fishes}, eds. J. S. Albert and R. E. Reis (University of California Press), 89--104.

\leavevmode\hypertarget{ref-albright_rules_2003}{}%
Albright, A., and Hayes, B. (2003). Rules vs. Analogy in English past tenses: A computational/experimental study. \emph{Cognition} 90, 119--161. doi:\href{https://doi.org/10.1016/S0010-0277(03)00146-X}{10.1016/S0010-0277(03)00146-X}.

\leavevmode\hypertarget{ref-baayen_analyzing_2008}{}%
Baayen, H. (2008). \emph{Analyzing linguistic data: A practical introduction to statistics using R}. Cambridge: Cambridge University Press.

\leavevmode\hypertarget{ref-baayen_word_2001}{}%
Baayen, R. H. (2001). \emph{Word frequency distributions}. Dordrecht, Netherlands: Springer Science+Business Media.

\leavevmode\hypertarget{ref-baker_word_2014}{}%
Baker, B. (2014). ``Word structure in Australian languages,'' in \emph{The languages and linguistics of Australia: A comprehensive guide} (Walter de Gruyter), 139--214.

\leavevmode\hypertarget{ref-barford_changes_1999}{}%
Barford, P., Bestavros, A., Bradley, A., and Crovella, M. (1999). Changes in web client access patterns: Characteristics and caching implications. \emph{World Wide Web} 2, 15--28. doi:\href{https://doi.org/10.1023/A:1019236319752}{10.1023/A:1019236319752}.

\leavevmode\hypertarget{ref-barford_generating_1998}{}%
Barford, P., and Crovella, M. (1998). Generating representative web workloads for network and server performance evaluation. in \emph{Measurement and modeling of computer systems (SIGMETRICS)} (New York: Association for Computing Machinery (ACM)), 151--160. doi:\href{https://doi.org/10.1145/277851.277897}{10.1145/277851.277897}.

\leavevmode\hypertarget{ref-barndorff-nielsen_inference_1994}{}%
Barndorff-Nielsen, O. E., and Cox, D. R. (1994). \emph{Inference and asymptotics}. London: Chapman \& Hall.

\leavevmode\hypertarget{ref-becker-kristal_acoustic_2010}{}%
Becker-Kristal, R. (2010). Acoustic typology of vowel inventories and Dispersion Theory: Insights from a large cross-linguistic corpus.

\leavevmode\hypertarget{ref-blasi_soundmeaning_2016}{}%
Blasi, D. E., Wichmann, S., Hammarström, H., Stadler, P. F., and Christiansen, M. H. (2016). Sound--meaning association biases evidenced across thousands of languages. \emph{PNAS} 113, 10818--10823. doi:\href{https://doi.org/10.1073/pnas.1605782113}{10.1073/pnas.1605782113}.

\leavevmode\hypertarget{ref-bloomfield_language_1935}{}%
Bloomfield, L. (1935). \emph{Language}. London: Allen \& Unwin.

\leavevmode\hypertarget{ref-boretzky1991contact}{}%
Boretzky, N. (1991). Contact-induced sound change. \emph{Diachronica} 8, 1--15.

\leavevmode\hypertarget{ref-borodovsky_general_1989}{}%
Borodovsky, M. Y., and Gusein-Zade, S. M. (1989). A general rule for ranged series of codon frequencies in different genomes. \emph{Journal of Biomolecular Structure and Dynamics} 6, 1001--1012. doi:\href{https://doi.org/10.1080/07391102.1989.10506527}{10.1080/07391102.1989.10506527}.

\leavevmode\hypertarget{ref-bowern_chirila:_2016}{}%
Bowern, C. (2016). Chirila: Contemporary and historical resources for the Indigenous languages of Australia. \emph{Language Documentation and Conservation} 10.

\leavevmode\hypertarget{ref-browman_towards_1986}{}%
Browman, C. P., and Goldstein, L. M. (1986). Towards an articulatory phonology. \emph{Phonology} 3, 219--252. Available at: \url{http://journals.cambridge.org/article_S0952675700000658}.

\leavevmode\hypertarget{ref-brzezinski_power_2014}{}%
Brzezinski, M. (2014). Power laws in citation distributions: Evidence from Scopus. \emph{arXiv e-prints}.

\leavevmode\hypertarget{ref-busby_distribution_1982}{}%
Busby, P. A. (1982). \emph{The distribution of phonemes in Australian Aboriginal languages}. Canberra: Pacific Linguistics.

\leavevmode\hypertarget{ref-capell_new_1956}{}%
Capell, A. (1956). \emph{A new approach to Australian linguistics}. Sydney: University of Sydney for Oceania.

\leavevmode\hypertarget{ref-chao_non-uniqueness_1934}{}%
Chao, Y.-R. (1934). The non-uniqueness of phonemic solutions of phonetic systems. \emph{Bulletin of the Institute of History and Philology, Academia Sinica} 4, 363--397.

\leavevmode\hypertarget{ref-cho_friendship_2011}{}%
Cho, E., Myers, S. A., and Leskovec, J. (2011). Friendship and mobility: User movement in location-based social networks. in \emph{Knowledge discovery and data mining (SIGKDD)} (New York: Association for Computing Machinery (ACM)), 1082--1090. doi:\href{https://doi.org/10.1145/2020408.2020579}{10.1145/2020408.2020579}.

\leavevmode\hypertarget{ref-chung_stochastic_1994}{}%
Chung, K. H., and Cox, R. A. K. (1994). A stochastic model of superstardom: An application of the yule distribution. \emph{The Review of Economics and Statistics}, 771--775.

\leavevmode\hypertarget{ref-clauset_power-law_2009}{}%
Clauset, A., Shalizi, C. R., and Newman, M. E. J. (2009). Power-law distributions in empirical data. \emph{SIAM Review} 51, 661--703. doi:\href{https://doi.org/10.1137/070710111}{10.1137/070710111}.

\leavevmode\hypertarget{ref-coleman_stochastic_1997}{}%
Coleman, J., and Pierrehumbert, J. (1997). Stochastic phonological grammars and acceptability. in \emph{ACL special interest group in computational phonology} (Somerset, NJ: Association for Computational Linguistics), 49--56. Available at: \url{http://arxiv.org/abs/cmp-lg/9707017}.

\leavevmode\hypertarget{ref-crawford_adaptation_2009}{}%
Crawford, C. J. (2009). Adaptation and transmission in Japanese loanword phonology. Available at: \url{http://core.ac.uk/download/pdf/4912071.pdf}.

\leavevmode\hypertarget{ref-cysouw_probability_2009}{}%
Cysouw, M. (2009). ``On the probability distribution of typological frequencies,'' in \emph{The mathematics of language}, eds. C. Ebert, G. Jäger, and J. Michaelis (Berlin: Springer), 29--35.

\leavevmode\hypertarget{ref-dixon_languages_1980}{}%
Dixon, R. M. W. (1980). \emph{The languages of Australia}. Cambridge: Cambridge University Press.

\leavevmode\hypertarget{ref-dockum_swadesh_2019}{}%
Dockum, R., and Bowern, C. (2019). Swadesh lists are not long enough: Drawing phonological generalizations from limited data. \emph{Language Documentation and Description} 16, 35--54.

\leavevmode\hypertarget{ref-downey_structural_2001}{}%
Downey, A. B. (2001). The structural cause of file size distributions. in \emph{Modeling, analysis and simulation of computer and telecommunication systems (MASCOTS)}, 361--370. doi:\href{https://doi.org/10.1109/MASCOT.2001.948888}{10.1109/MASCOT.2001.948888}.

\leavevmode\hypertarget{ref-dresher_contrastive_2009}{}%
Dresher, B. E. (2009). \emph{The contrastive hierarchy in phonology}. Cambridge: Cambridge University Press.

\leavevmode\hypertarget{ref-eddington_spanish_2004}{}%
Eddington, D. (2004). \emph{Spanish phonology and morphology: Experimental and quantitative perspectives}. Amsterdam: John Benjamins.

\leavevmode\hypertarget{ref-eeckhout_gibrats_2004}{}%
Eeckhout, J. (2004). Gibrat's law for (all) cities. \emph{American Economic Review} 94, 1429--1451. doi:\href{https://doi.org/10.1257/0002828043052303}{10.1257/0002828043052303}.

\leavevmode\hypertarget{ref-ernestus_predicting_2003}{}%
Ernestus, M. T. C., and Baayen, R. H. (2003). Predicting the unpredictable: Interpreting neutralized segments in dutch. \emph{Language} 79, 5--38. doi:\href{https://doi.org/10.1353/lan.2003.0076}{10.1353/lan.2003.0076}.

\leavevmode\hypertarget{ref-estoup_gammes_1916}{}%
Estoup, J. B. (1916). \emph{Gammes Sténographiques: Méthode et exercices pour l'acquisition de la vitesse}. Institut Sténographique.

\leavevmode\hypertarget{ref-everett_similar_2018}{}%
Everett, C. (2018a). The similar rates of occurrence of consonants across the world's languages: A quantitative analysis of phonetically transcribed word lists. \emph{Language Sciences} 69, 125--135. doi:\href{https://doi.org/10.1016/j.langsci.2018.07.003}{10.1016/j.langsci.2018.07.003}.

\leavevmode\hypertarget{ref-everett_climate_2015}{}%
Everett, C., Blasi, D. E., and Roberts, S. G. (2015). Climate, vocal folds, and tonal languages: Connecting the physiological and geographic dots. \emph{Proceedings of the National Academy of Sciences} 112, 1322--1327. doi:\href{https://doi.org/10.1073/pnas.1417413112}{10.1073/pnas.1417413112}.

\leavevmode\hypertarget{ref-everett_global_2018}{}%
Everett, C. D. (2018b). The global dispreference for posterior voiced obstruents: A quantitative assessment of word-list data. \emph{Language} 94, e311--e323.

\leavevmode\hypertarget{ref-faloutsos_power-law_1999}{}%
Faloutsos, M., Faloutsos, P., and Faloutsos, C. (1999). On power-law relationships of the internet topology. in \emph{Applications, technologies, architectures, and protocols for computer communication (SIGCOMM)} (New York: Association for Computing Machinery (ACM)), 251--262. doi:\href{https://doi.org/10.1145/316188.316229}{10.1145/316188.316229}.

\leavevmode\hypertarget{ref-farrell_comprehensive_2006}{}%
Farrell, P. J., and Rogers-Stewart, K. (2006). Comprehensive study of tests for normality and symmetry: Extending the Spiegelhalter test. \emph{Journal of Statistical Computation and Simulation} 76, 803--816. doi:\href{https://doi.org/10.1080/10629360500109023}{10.1080/10629360500109023}.

\leavevmode\hypertarget{ref-flege1987production}{}%
Flege, J. E. (1987). The production of ``new'' and ``similar'' phones in a foreign language: Evidence for the effect of equivalence classification. \emph{Journal of phonetics} 15, 47--65.

\leavevmode\hypertarget{ref-frank_how_2014}{}%
Frank, S. A. (2014). How to read probability distributions as statements about process. \emph{Entropy} 16, 6059--6098. doi:\href{https://doi.org/10.3390/e16116059}{10.3390/e16116059}.

\leavevmode\hypertarget{ref-gabaix_zipfs_1999}{}%
Gabaix, X. (1999). Zipf's law for cities: An explanation. \emph{The Quarterly Journal of Economics} 114, 739--767. doi:\href{https://doi.org/10.1162/003355399556133}{10.1162/003355399556133}.

\leavevmode\hypertarget{ref-gasser_revisiting_2014}{}%
Gasser, E., and Bowern, C. (2014). Revisiting phonotactic generalizations in Australian languages. \emph{Annual Meeting on Phonology AMP} 1. doi:\href{https://doi.org/10.3765/amp.v1i1.17}{10.3765/amp.v1i1.17}.

\leavevmode\hypertarget{ref-gillespie_fitting_2014}{}%
Gillespie, C. S. (2014). Fitting heavy tailed distributions: The poweRlaw package. \emph{arXiv e-prints}.

\leavevmode\hypertarget{ref-good_statistics_1969}{}%
Good, I. J. (1969). ``Statistics of language,'' in \emph{Encyclopaedia of linguistics, information, and control}, eds. A. R. Meetham and R. A. Hudson (Oxford: Pergamon Press).

\leavevmode\hypertarget{ref-gusein-zade_distribution_1988}{}%
Gusein-Zade, S. M. (1988). On the distribution of letters of the Russian language by frequencies. \emph{Problems of the Transmission of Information} 23, 103--107.

\leavevmode\hypertarget{ref-hamilton_phonetic_1996}{}%
Hamilton, P. J. (1996). Phonetic constraints and markedness in the phonotactics of Australian Aboriginal languages.

\leavevmode\hypertarget{ref-hayes_stochastic_2006}{}%
Hayes, B., and Londe, Z. C. (2006). Stochastic phonological knowledge: The case of hungarian vowel harmony. \emph{Phonology} 23, 59--104. doi:\href{https://doi.org/10.1017/S0952675706000765}{10.1017/S0952675706000765}.

\leavevmode\hypertarget{ref-hockett_problem_1963}{}%
Hockett, C. F. (1963). ``The problem of universals in language,'' in \emph{Universals of language}, ed. J. Greenberg (Cambridge, MA: MIT Press), 1--29.

\leavevmode\hypertarget{ref-hoenigswald1965language}{}%
Hoenigswald, H. M. (1965). Language change and linguistic reconstruction.

\leavevmode\hypertarget{ref-hudson_walmajarri_1993}{}%
Hudson, J., and Richards, E. (1993). Walmajarri dictionary. Available at: \url{aseda.aiatsis.gov.au}.

\leavevmode\hypertarget{ref-van_der_hulst_phonological_2017}{}%
Hulst, H. van der (2017). ``Phonological typology,'' in \emph{The Cambridge handbook of linguistic typology}, eds. A. Y. Aikhenvald and R. M. W. Dixon (Cambridge: Cambridge University Press), 39--77.

\leavevmode\hypertarget{ref-hyman_universals_2008}{}%
Hyman, L. M. (2008). Universals in phonology. \emph{The Linguistic Review} 25, 83--137. doi:\href{https://doi.org/10.1515/TLIR.2008.003}{10.1515/TLIR.2008.003}.

\leavevmode\hypertarget{ref-johnson_individual_1993}{}%
Johnson, K., Ladefoged, P., and Lindau, M. (1993). Individual differences in vowel production. \emph{The Journal of the Acoustical Society of America} 94, 701--714.

\leavevmode\hypertarget{ref-kang2003perceptual}{}%
Kang, Y. (2003). Perceptual similarity in loanword adaptation: English postvocalic word-final stops in korean. \emph{Phonology}, 219--273.

\leavevmode\hypertarget{ref-kiparsky_formal_2018}{}%
Kiparsky, P. (2018). ``Formal and empirical issues in phonological typology,'' in \emph{Phonological typology}, eds. L. M. Hyman and F. Plank (Berlin: De Gruyter Mouton), 54--106.

\leavevmode\hypertarget{ref-kuba2012limiting}{}%
Kuba, M., and Panholzer, A. (2012). Limiting distributions for a class of diminishing urn models. \emph{Advances in Applied Probability} 44, 87--116.

\leavevmode\hypertarget{ref-kucera_computational_1967}{}%
Kucera, H., Francis, W. N., Carroll, J. B., and Twaddell, W. F. (1967). \emph{Computational analysis of present day American English}. 1st ed. Providence, RI: Brown University Press.

\leavevmode\hypertarget{ref-kullback_information_1951}{}%
Kullback, S., and Leibler, R. A. (1951). On information and sufficiency. \emph{The Annals of Mathematical Statistics} 22, 79--86.

\leavevmode\hypertarget{ref-lass_vowel_1984}{}%
Lass, R. (1984). Vowel system universals and typology: Prologue to theory. \emph{Phonology Yearbook} 1, 75--111. doi:\href{https://doi.org/10.1017/S0952675700000300}{10.1017/S0952675700000300}.

\leavevmode\hypertarget{ref-lee_change_2018}{}%
Lee, Y., and Kim, I. (2018). Change and stability in shopping tourist destination networks: The case of Seoul in Korea. \emph{Journal of Destination Marketing \& Management}. doi:\href{https://doi.org/10.1016/j.jdmm.2018.02.004}{10.1016/j.jdmm.2018.02.004}.

\leavevmode\hypertarget{ref-lev2014effect}{}%
Lev-Ari, S., San Giacomo, M., and Peperkamp, S. (2014). The effect of domain prestige and interlocutors' bilingualism on loanword adaptations. \emph{Journal of Sociolinguistics} 18, 658--684.

\leavevmode\hypertarget{ref-levy_gibrats_2009}{}%
Levy, M. (2009). Gibrat's law for (all) cities: Comment. \emph{American Economic Review} 99, 1672--1675. doi:\href{https://doi.org/10.1257/aer.99.4.1672}{10.1257/aer.99.4.1672}.

\leavevmode\hypertarget{ref-li_random_1992}{}%
Li, W. (1992). Random texts exhibit Zipf's-law-like word frequency distribution. \emph{IEEE Transactions on Information Theory} 38, 1842--1845. doi:\href{https://doi.org/10.1109/18.165464}{10.1109/18.165464}.

\leavevmode\hypertarget{ref-liljencrants_numerical_1972}{}%
Liljencrants, J., and Lindblom, B. (1972). Numerical simulation of vowel quality systems: The role of perceptual contrast. \emph{Language} 48, 839--862.

\leavevmode\hypertarget{ref-loeb1989formal}{}%
Loeb, D. E., and Rota, G.-C. (1989). Formal power series of logarithmic type. \emph{Advances in Mathematics} 75, 1--118.

\leavevmode\hypertarget{ref-macklin-cordes_phylogenetic_2020}{}%
Macklin-Cordes, J. L., Bowern, C., and Round, E. R. (Accepted). Phylogenetic signal in phonotactics. \emph{Diachronica}.

\leavevmode\hypertarget{ref-macklin-cordes_high-definition_2015}{}%
Macklin-Cordes, J. L., and Round, E. R. (2015). High-definition phonotactics reflect linguistic pasts. in \emph{Quantitative investigations in theoretical linguistics (QITL)}, eds. J. Wahle, M. Köllner, H. Baayen, G. Jäger, and T. Baayen-Oudshoorn (Tübingen: University of Tübingen). doi:\href{https://doi.org/10.15496/publikation-8609}{10.15496/publikation-8609}.

\leavevmode\hypertarget{ref-maddieson_patterns_1984}{}%
Maddieson, I. (1984). \emph{Patterns of sounds}. Cambridge: Cambridge University Press.

\leavevmode\hypertarget{ref-malevergne_testing_2011}{}%
Malevergne, Y., Pisarenko, V., and Sornette, D. (2011). Testing the Pareto against the lognormal distributions with the uniformly most powerful unbiased test applied to the distribution of cities. \emph{Physical Review E} 83, 036111. doi:\href{https://doi.org/10.1103/PhysRevE.83.036111}{10.1103/PhysRevE.83.036111}.

\leavevmode\hypertarget{ref-mandelbrot_structure_1954}{}%
Mandelbrot, B. (1954). Structure formelle des textes et communication. \emph{WORD} 10, 1--27. doi:\href{https://doi.org/10.1080/00437956.1954.11659509}{10.1080/00437956.1954.11659509}.

\leavevmode\hypertarget{ref-martindale_comparison_1996}{}%
Martindale, C., Gusein-Zade, S. M., McKenzie, D., and Borodovsky, M. Y. (1996). Comparison of equations describing the ranked frequency distributions of graphemes and phonemes. \emph{Journal of Quantitative Linguistics} 3, 106--112. doi:\href{https://doi.org/10.1080/09296179608599620}{10.1080/09296179608599620}.

\leavevmode\hypertarget{ref-mitzenmacher_brief_2004}{}%
Mitzenmacher, M. (2004). A brief history of generative models for power law and lognormal distributions. \emph{Internet Mathematics} 1, 226--251. doi:\href{https://doi.org/10.1080/15427951.2004.10129088}{10.1080/15427951.2004.10129088}.

\leavevmode\hypertarget{ref-montemurro_beyond_2001}{}%
Montemurro, M. A. (2001). Beyond the Zipf--Mandelbrot law in quantitative linguistics. \emph{Physica A: Statistical Mechanics and its Applications} 300, 567--578. doi:\href{https://doi.org/10.1016/S0378-4371(01)00355-7}{10.1016/S0378-4371(01)00355-7}.

\leavevmode\hypertarget{ref-moran_phonetics_2012}{}%
Moran, S. (2012). Phonetics information base and lexicon.

\leavevmode\hypertarget{ref-moran_unicode_2018}{}%
Moran, S., and Cysouw, M. (2018). \emph{The Unicode cookbook for linguists: Managing writing systems using orthography profiles}. Berlin: Language Science Press.

\leavevmode\hypertarget{ref-evolang12}{}%
Moran, S., and Verkerk, A. (2018). Differential rates of change in consonant and vowel systems. in \emph{The evolution of language: Proceedings of the 12th international conference (evolangxii)}, eds. C. Cuskley, M. Flaherty, H. Little, L. McCrohon, A. Ravignani, and T. Verhoef (NCU Press). doi:\href{https://doi.org/10.12775/3991-1.077}{10.12775/3991-1.077}.

\leavevmode\hypertarget{ref-naranan_information_1992}{}%
Naranan, S., and Balasubrahmanyan, V. K. (1992). Information theoretic models in statistical linguistics---part i: A model for word frequencies. \emph{Current Science} 63, 261--269.

\leavevmode\hypertarget{ref-naranan_models_1998}{}%
Naranan, S., and Balasubrahmanyan, V. K. (1998). Models for power law relations in linguistics and information science. \emph{Journal of Quantitative Linguistics} 5, 35--61. doi:\href{https://doi.org/10.1080/09296179808590110}{10.1080/09296179808590110}.

\leavevmode\hypertarget{ref-newman_power_2005}{}%
Newman, M. E. J. (2005). Power laws, Pareto distributions and Zipf's law. \emph{Contemporary Physics} 46, 323--351. doi:\href{https://doi.org/10.1080/00107510500052444}{10.1080/00107510500052444}.

\leavevmode\hypertarget{ref-paradis1997preservation}{}%
Paradis, C., and LaCharité, D. (1997). Preservation and minimality in loanword adaptation. \emph{Journal of linguistics}, 379--430.

\leavevmode\hypertarget{ref-pareto_cours_1897}{}%
Pareto, V. (1897). \emph{Cours d'Economie politique}. Lausanne, France: F. Rouge.

\leavevmode\hypertarget{ref-piantadosi2014zipf}{}%
Piantadosi, S. T. (2014). Zipf's word frequency law in natural language: A critical review and future directions. \emph{Psychonomic bulletin \& review} 21, 1112--1130.

\leavevmode\hypertarget{ref-proctor_gestural_2009}{}%
Proctor, M. I. (2009). Gestural characterization of a phonological class: The liquids.

\leavevmode\hypertarget{ref-r-core-team_r_2017}{}%
R Core Team (2017). \emph{R: A language and environment for statistical computing}. Vienna, Austria: R Foundation for Statistical Computing Available at: \url{https://www.R-project.org/}.

\leavevmode\hypertarget{ref-round_matthew_2017}{}%
Round, E. R. (2017a). Matthew K. Gordon: Phonological typology (review). \emph{Folia Linguistica} 51, 745--755. doi:\href{https://doi.org/10.1515/flin-2017-0027}{10.1515/flin-2017-0027}.

\leavevmode\hypertarget{ref-round_phonemic_2019}{}%
Round, E. R. (2019). ``Phonemic inventories of Australia (database of 392 languages),'' in \emph{PHOIBLE 2.0}, eds. S. Moran and D. McCloy (Jena, Germany: Max Planck Institute for the Science of Human History).

\leavevmode\hypertarget{ref-round_segment_2020}{}%
Round, E. R. (2020). ``Segment inventories in Australian languages,'' in \emph{Oxford guide to Australian languages}, ed. C. Bowern (Oxford: Oxford University Press).

\leavevmode\hypertarget{ref-round_ausphon-lexicon_2017}{}%
Round, E. R. (2017b). The AusPhon-Lexicon project: 2 million normalized segments across 300 Australian languages. in \emph{Poznań linguistic meeting} (Poznań, Poland). Available at: \url{http://wa.amu.edu.pl/plm_old/2017/files/abstracts/PLM2017_Abstract_Round.pdf}.

\leavevmode\hypertarget{ref-sigurd_rank-frequency_1968}{}%
Sigurd, B. (1968). Rank-frequency distributions for phonemes. \emph{Phonetica} 18, 1--15. doi:\href{https://doi.org/10.1159/000258595}{10.1159/000258595}.

\leavevmode\hypertarget{ref-simon_class_1955}{}%
Simon, H. A. (1955). On a class of skew distribution functions. \emph{Biometrika} 42, 425--440. doi:\href{https://doi.org/10.1093/biomet/42.3-4.425}{10.1093/biomet/42.3-4.425}.

\leavevmode\hypertarget{ref-spiegelhalter_omnibus_1980}{}%
Spiegelhalter, D. J. (1980). An omnibus test for normality for small samples. \emph{Biometrika} 67, 493--496. doi:\href{https://doi.org/10.1093/biomet/67.2.493}{10.1093/biomet/67.2.493}.

\leavevmode\hypertarget{ref-stevens_quantal_1989}{}%
Stevens, K. N. (1989). On the quantal nature of speech. \emph{Journal of phonetics} 17, 3--45.

\leavevmode\hypertarget{ref-stumpf_critical_2012}{}%
Stumpf, M. P. H., and Porter, M. A. (2012). Critical truths about power laws. \emph{Science} 335, 665--666. doi:\href{https://doi.org/10.1126/science.1216142}{10.1126/science.1216142}.

\leavevmode\hypertarget{ref-tambovtsev_phoneme_2007}{}%
Tambovtsev, Y., and Martindale, C. (2007). Phoneme frequencies follow a Yule distribution. \emph{SKASE Journal of Theoretical Linguistics} 4.

\leavevmode\hypertarget{ref-touboul_can_2010}{}%
Touboul, J., and Destexhe, A. (2010). Can power-law scaling and neuronal avalanches arise from stochastic dynamics? \emph{PLOS ONE} 5, e8982. doi:\href{https://doi.org/10.1371/journal.pone.0008982}{10.1371/journal.pone.0008982}.

\leavevmode\hypertarget{ref-urzua_testing_2011}{}%
Urzúa, C. M. (2011). Testing for Zipf's law: A common pitfall. \emph{Economics Letters} 112, 254--255. doi:\href{https://doi.org/10.1016/j.econlet.2011.05.049}{10.1016/j.econlet.2011.05.049}.

\leavevmode\hypertarget{ref-vuong_likelihood_1989}{}%
Vuong, Q. H. (1989). Likelihood ratio tests for model selection and non-nested hypotheses. \emph{Econometrica} 57, 307--333. doi:\href{https://doi.org/10.2307/1912557}{10.2307/1912557}.

\leavevmode\hypertarget{ref-whitworth_choice_1901}{}%
Whitworth, W. A. (1901). \emph{Choice and chance: With 1,000 exercises}. Cambridge: Deighton Bell; Company.

\leavevmode\hypertarget{ref-witten_source_1990}{}%
Witten, I. H., and Bell, T. C. (1990). Source models for natural language text. \emph{International Journal of Man-Machine Studies} 32, 545--579. doi:\href{https://doi.org/10.1016/S0020-7373(05)80033-1}{10.1016/S0020-7373(05)80033-1}.

\leavevmode\hypertarget{ref-yule_mathematical_1925}{}%
Yule, G. U. (1925). A mathematical theory of evolution, based on the conclusions of Dr. J. C. Willis, F. R. S. \emph{Phil. Trans. R. Soc. Lond. B} 213, 21--87. doi:\href{https://doi.org/10.1098/rstb.1925.0002}{10.1098/rstb.1925.0002}.

\leavevmode\hypertarget{ref-zipf_human_1949}{}%
Zipf, G. K. (1949). \emph{Human behavior and the principle of least effort}. Reading, MA: Addison-Wesley.

\leavevmode\hypertarget{ref-zipf_selective_1932}{}%
Zipf, G. K. (1932). \emph{Selective studies and the principle of relative frequency in language}. Cambridge, MA: Harvard University Press.

\leavevmode\hypertarget{ref-zuraw_patterned_2000}{}%
Zuraw, K. R. (2000). Patterned exceptions in phonology.

\endgroup

\endgroup

\clearpage

\hypertarget{tables-and-figures}{%
\section*{Tables and Figures}\label{tables-and-figures}}
\addcontentsline{toc}{section}{Tables and Figures}

\begin{table}[!h]

\caption{\label{tab:pl-summary}Power law (without $x_{min}$). Summary of $\alpha$ paramter, goodness-of-fit and $p$ values for the power law distribution fitted to each language's full phonemic inventory.}
\centering
\begin{tabular}[t]{lcccc}
\toprule
\textbf{ } & \textbf{Mean} & \textbf{SD} & \textbf{Min} & \textbf{Max}\\
\midrule
$\alpha$ & 1.38 & 0.17 & 1.16 & 2.18\\
goodness-of-fit & 0.35 & 0.07 & 0.15 & 0.53\\
$p$ & 0.01 & 0.03 & 0.00 & 0.27\\
\bottomrule
\end{tabular}
\end{table}

\begin{table}[!h]

\caption{\label{tab:pl-xmin-summary}Power law distribution (with $x_{min}$). Summary of $\alpha$ paramter, goodness-of-fit and $p$ values for the power law distribution fitted to a subset of more frequent phonemes in each language.}
\centering
\begin{tabular}[t]{lcccc}
\toprule
\textbf{ } & \textbf{Mean} & \textbf{SD} & \textbf{Min} & \textbf{Max}\\
\midrule
$\alpha$ & 2.75 & 0.65 & 1.51 & 6.14\\
goodness-of-fit & 0.14 & 0.03 & 0.08 & 0.22\\
$p$ & 0.62 & 0.26 & 0.01 & 1.00\\
\bottomrule
\end{tabular}
\end{table}

\begin{table}[!h]

\caption{\label{tab:results-summary}Results summary. For each of the four distributions considered, this table lists the number of languages (and percentage of the total language sample) for which the distribution plausibly fits, as indicated by an uncorrected $p > 0.1$ value. 'Prop. fitted' gives the average proportion of each language's phoneme inventory above $x_{min}$.}
\centering
\begin{tabular}[t]{lccc}
\toprule
\textbf{ } & \textbf{Without $x_{min}$} & \textbf{With $x_{min}$} & \textbf{Prop. fitted}\\
\midrule
Power law & 2 (1\%) & 158 (95\%) & 56\%\\
Lognormal & 93 (56\%) & 155 (93\%) & 78\%\\
Exponential & 147 (89\%) & 146 (88\%) & 84\%\\
Poisson & 0 (0\%) & 43 (26\%) & 17\%\\
\bottomrule
\end{tabular}
\end{table}

\end{document}